\documentclass[10pt]{article} 
\usepackage[preprint]{tmlr}

\usepackage{amsmath,amsfonts,bm}

\def\eqref#1{equation~\ref{#1}}

\def\1{\bm{1}}

\DeclareMathAlphabet{\mathsfit}{\encodingdefault}{\sfdefault}{m}{sl}
\SetMathAlphabet{\mathsfit}{bold}{\encodingdefault}{\sfdefault}{bx}{n}

\newcommand{\R}{\mathbb{R}}

\usepackage{hyperref}
\usepackage{url}

\usepackage{algorithm}
\usepackage{algpseudocode}
\usepackage{siunitx}
\usepackage{graphicx}
\usepackage{booktabs}

\title{Steering Neural Network Training through Interpretable Constraints Based on Partial Dependence}

\author{\name Yann Claes \email y.claes@uliege.be \\
      \addr Montefiore Institute\\
      University of Liège
      \AND
      \name Pierre Geurts \\
      \addr Montefiore Institute\\
      University of Liège
      \AND
      \name V\^an Anh Huynh-Thu \\
      \addr Montefiore Institute\\
      University of Liège}

\newtheorem{definition}{Definition}
\newcommand{\x}{\mathbf{x}}
\newcommand{\N}{\mathbb{N}}

\def\month{MM}  
\def\year{YYYY} 
\def\openreview{\url{https://openreview.net/forum?id=XXXX}} 

\begin{document}

\maketitle

\begin{abstract}
Over the last few years, there has been an increased interest in making machine learning models more interpretable. Although a great deal of effort goes into developing techniques for interpreting the interactions learned by a given model, fewer studies focus on assessing the quality of such explanations. Even fewer focus on how to adjust the model to produce explanations faithful to prior knowledge, a process known as explanation-guided learning. Furthermore, most approaches in this area focus on classification problems and usually assume prior knowledge about which input features or regions are most important. In this work, we introduce a new approach to steering neural networks based on partial dependence, such that their average response to certain features aligns with specific functional domain knowledge about the problem. We empirically demonstrate on a range of regression problems, including dynamical systems forecasting, that models whose training has been controlled using our method perform better than unconstrained models and are more data-efficient. Moreover, we highlight that interpretations obtained from the former actually align with the user-provided knowledge, whereas those obtained from the latter do not.
\end{abstract}

\section{Introduction}
Over the past decades, progress in machine learning (ML) methods has enabled practitioners in a variety of fields to tackle increasingly complex problems, in which such models serve as surrogates for first-principles models or costly simulators. While they come with great expressiveness, it is nevertheless not straightforward to interpret their reasoning process, which is why they are often referred to as "black box" models.
There has been recently a vast development of the field of explainable artificial intelligence (XAI), in a desire to increase our understanding of decisions made by ML models \citep{adadi2018peeking,gilpin2018explaining,linardatos2020explainable}. 

On the one hand, models can be made more interpretable, through specific design choices. For instance, \citet{nelder1972generalized} introduced generalized linear models (GLM) to model various response distributions, which are more flexible than linear regression, yet still very simple to explain. \citet{hastie1986generalized} later implemented generalized additive models (GAM) as an extension to GLMs with non-linear predictors, and \citet{caruana2015intelligible} successfully applied GAMs to healthcare problems. 

On the other hand, model predictions can be interpreted through model-specific and model-agnostic interpretation tools, with applications in different domains such as image processing \citep{zhang2018visual}, healthcare \citep{vellido2020importance} and neural language processing \citep{madsen2022post}. For example, \citet{lundberg2017unified} introduced SHAP values to compute feature importance in the prediction of the model. In a similar philosophy, \citet{friedman2001greedy} designed partial dependence plots to visualize the marginal contribution of certain features to the model response, and complementary plotting functions were later implemented \citep{goldstein2015peeking,apley2020visualizing}. For interpretations of visual predictions, most works rely on class activation mapping techniques to localize regions of importance \citep{zhou2016learning,selvaraju2017grad,chattopadhay2018grad}.

Although current interpretation tools can help domain experts to understand how decisions are made, they cannot take their feedback into account, which could be important if the model provides a correct decision for the wrong reasons. \citet{gao2024going} notice that most works in XAI focus on the generation of model explanations, but leave aside the assessment of their quality, and judiciously state that adjusting models to produce more reasonable explanations is as important as producing them. For that purpose, several works have developed approaches such that, after training, model explanations align with specific domain knowledge \citep{zhang2016rationale,zhang2024megl}, a process known as explanation-guided learning (EGL). Furthermore, previous works in EGL have demonstrated that leveraging explanation knowledge could also help with limited training sizes \citep{gao2022res,gao2022aligning} and with out-of-distribution data \citep{taesiri2022visual}.

In this work, we introduce a novel approach to control neural network training, based on partial dependence, such that its marginal response to certain inputs aligns with specific knowledge about the problem. We empirically validate the performance of our algorithm on several regression problems (including dynamical systems forecasting) and show that the obtained models perform better than their unconstrained counterparts. Moreover, they produce explanations faithful to prior knowledge, require less samples for training, and generalize better outside the originally considered input domain. 

\section{Related work}

Explanation-guided learning is an emerging research field whose goal is to steer the training process of ML models to improve their explainability, according to specific domain knowledge. Several works implement these principles, mostly for knowledge in the form of importance of input features and input regions \citep{rieger2020interpretations,zhang2024megl}, or in the form of global properties that the model should satisfy, such as sparsity \citep{wu2020regional}, or even to mitigate bias \citep{liu2019incorporatingpriorsfeatureattribution}. In comparison with previous works, our approach rather assumes knowledge in the form of a specific function of a subset of features, i.e. the partial dependence. Therefore, our approach is not directly comparable to previous works since the given knowledge forms are fundamentally different. Furthermore, while previous works focus on classification tasks based on images or textual data, our approach rather focuses on regression tasks applied with tabular data or dynamical systems.

ML models can be trained specifically to satisfy known constraints about the problem. For example, convexity \citep{pya2015shape,amos2017input} and monotonicity \citep{pya2015shape,wehenkel2019unconstrained} constraints can be enforced on neural networks or other types of models. Structural constraints can also be required directly on the output of the models \citep{hendriks2020linearly,chen2023structured}. Nevertheless, for all these approaches, constraints are imposed on the function, not on the explanations.

Another related line of work is that of physics-informed machine learning. In this research area, physics-based models are combined with ML models to enhance physical consistency and improve generalization. In practice, most hybrid models sum the contributions of the first-principles model and of the ML model \citep{qian2021integrating,takeishi2021physics}. This decomposition implies finding the right balance between the physics-based and ML models, which has been investigated in several works \citep{yin2021augmenting,dona2022constrained,claes2025hybrid}. In opposition to such approaches, our method is less restrictive as it does not assume additivity or any other form of model hybridization, but simply relies on a single ML model. Furthermore, these training algorithms do not take into account the explanations of the resulting models, hence frameworks are radically different.

\section{Methodology}
\label{sec:methodology}
In this section, we first provide some mathematical background and intuition about the partial dependence as a model explanation. We then formally define the class of problems that is studied, along with our proposed training algorithm and the different configurations that we consider. We focus here on standard regression problems. Appendix \ref{app:subsec_dyn_systems} explains how the approach can be extended to dynamical systems forecasting.

\subsection{Partial dependence}
Let us define the regression problem $y = f(\x) + \varepsilon$, where $y \in \R$ and $\x \in \R^d$, with $d \in \N_+$, drawn from a distribution $p(\x,y)$ and $\varepsilon \sim \mathcal{N}(0, \sigma^2)$ some Gaussian noise.
\begin{definition}
    Let $\x_k$ be some subset of features of interest and $\x_{-k}$ its complement, with $\x_k \cup \x_{-k} = \x$, then the partial dependence (PD) of a function $f(\x)$ on $\x_k$ is defined as \citep{friedman2001greedy}:
    \begin{align}
    \label{eq:pdp_definition}
    \begin{split}
        \textit{PD}_{\infty}(f,\x_k) &= \mathbb{E}_{\x_{-k}}\left[f(\x_k, \x_{-k})\right],\\ 
        &= \int f(\x_k, \x_{-k}) p(\x_{-k}) d\x_{-k},
    \end{split}
    \end{align}
    with $p(\x_{-k})$ the marginal distribution of $\x_{-k}$.
\end{definition}

In practice, with a finite learning set $\textit{LS} = \{(\x^{(i)}, y^{(i)})\}_{i=1}^{N}$ of $N$ samples, the PD is estimated by
\begin{equation}
    \label{eq:sample_pd}
    \textit{PD}_N(f, \x_k) = \frac{1}{N}\sum_{i=1}^{N}f(\x_k, \x^{(i)}_{-k}).
\end{equation}

In other words, the PD measures how a certain subset of features $\x_k$ impacts the output of a function $f$, on average. Now, let us assume that the entire effect of $\x_k$ is known algebraically as $f_k(\x_k)$, some function of $\x_k$. If such effect is purely additive, i.e. $f(\x) = f_k(\x_k) + f_{-k}(\x_{-k})$, where $f_{-k}(\x_{-k})$ is an unknown function of  $\x_{-k}$, then 
\begin{equation*}
    \textit{PD}_{\infty}(f, \x_{k}) = f_k(\x_k) + \mathbb{E}_{\x_{-k}}\left[f_{-k}(\x_{-k})\right].
\end{equation*}
If the effect of $\x_k$ is instead multiplicative, i.e. $f(\x) = f_k(\x_k) \cdot f_{-k}(\x_{-k})$, then it follows
\begin{equation*}
    \textit{PD}_{\infty}(f, \x_{k}) = f_k(\x_k) \cdot \mathbb{E}_{\x_{-k}}\left[f_{-k}(\x_{-k})\right].
\end{equation*}

The PD is thus straightforward to compute in both cases, and we can write
\begin{equation}
    \textit{PD}_{\infty}(f, \x_k) = \phi_0^* \cdot f_k(\x_k) + \phi_1^*,
\end{equation}
where $\phi_0^* \in \R$ and $\phi_1^* \in \R$ are scaling and bias constants. To illustrate, let us consider the \textit{Friedman problem} \citep{friedman1983multidimensional}, which is used in our experiments below:
\begin{align*}
    f(\x) = \psi_{0} \sin(&\psi_{1} x_0 x_1) + \psi_{2} (x_2 - \psi_{3})^2 + \psi_{4} x_3 + \psi_{5} x_4.
\end{align*}
Taking $\x_k = \{x_0, x_1\}$, $\x_{-k} = \{x_2, x_3, x_4\}$, it follows
\begin{align*}
    \textit{PD}_{\infty}(f, \x_k) 
    &= \phi_0^* \cdot \psi_{0}\sin(\psi_{1} x_0 x_1) + \phi_1^*,
\end{align*}
where $\phi_0^* = 1$, $\phi_1^* = \mathbb{E}_{\x_{-k}}[\psi_{2} (x_2 - \psi_{3})^2 + \psi_{4} x_3 + \psi_{5} x_4]$.

\subsection{Problem statement}
We assume knowledge about model explanations in the form of a PD function, and are given a fixed learning set $\textit{LS} = \{(\x^{(i)}, y^{(i)})\}_{i=1}^{N}$ of $N$ samples.
Formally, let us assume some differentiable ML model $h^{\theta}(\x)$ where $\theta \in \R^{d_{\theta}}$ is a trainable parameter vector. We want to learn $\theta$ such that
\begin{align}
    \theta^* &= \arg\min_{\theta} \mathcal{L}_{\textit{tot}}(h^{\theta}, f; \textit{LS}) \\
     \text{s.t.} \quad \textit{PD}_N(h^{\theta}, \x_k) &= \textit{PD}_N(f, \x_k), \quad \forall \x_k \in \R^{d_{\x_k}} \label{eq:true_pd_constraint}
\end{align}
where
\begin{equation}
\label{eq:sample_mse_def}
    \mathcal{L}_{\textit{tot}}(h^{\theta}, f; \textit{LS}) = \frac{1}{N}\sum_{i=1}^{N}(h^{\theta}(\x^{(i)}) - y^{(i)})^2.
\end{equation}
In all generality, \eqref{eq:true_pd_constraint} assumes perfect knowledge of the PD up through the values of its parameters, which could be unrealistic in some cases. We hence assume that we do not have a perfect knowledge of the PD, but rather of the algebraic shape that the PD should have. Therefore, we introduce a function $h_k^{\phi}(\x_k)$, where $\phi \in \R^{d_{\phi}}$ is a learnable parameter vector including the previously defined scaling and bias parameters $\phi_0$ and $\phi_1$, such that for the optimal vector $\phi^*$, we have $h_k^{\phi^*}(\x_k) = \textit{PD}_{\infty}(f, \x_k)$. Taking back the previous example yields $h_k^{\phi}(\x_k) = \phi_0 \cdot [\phi_2\sin(\phi_3 x_0 x_1)] + \phi_1$, and $\phi^* = [1, C, \psi_0, \psi_1]$ would be a potential optimal vector, with $C = \mathbb{E}_{\x_{-k}}[\psi_{2} (x_2 - \psi_{3})^2 + \psi_{4} x_3 + \psi_{5} x_4]$. 

As \eqref{eq:true_pd_constraint} is hard to satisfy tightly, we rely on the following relaxation, introducing a coefficient $\lambda \in [0,1]$
\begin{align}
\label{eq:final_objective}
    \theta^*, \phi^* = \arg \min_{\theta, \phi} \mathcal{L}_{\lambda}(h^{\theta}, h_k^{\phi}, f; \textit{LS}),
\end{align}
where
\begin{equation*}
    \mathcal{L}_{\lambda}(h^{\theta}, h_k^{\phi}, f; \textit{LS}) = 
         (1-\lambda) \mathcal{L}_{\textit{tot}}(h^{\theta}, f; \textit{LS})
    + \lambda \mathcal{L}_{\textit{PD}}(h^{\theta}, h_k^{\phi}; \textit{LS}).
\end{equation*}
The second term is defined as

\begin{align}
\label{eq:sample_pd_mse}
    \mathcal{L}_{\textit{PD}}(h^{\theta}, h_k^{\phi}; \textit{LS})=
    \frac{1}{N}\sum_{i=1}^{N}&(\textit{PD}_N(h^{\theta}, \x_k^{(i)}) - h_k^{\phi}(\x_k^{(i)}))^2.
\end{align}

In practice, we apply steps of gradient descent to fit both $\theta$ and $\phi$ in \eqref{eq:final_objective}, for a certain number of epochs $E$. Although fitting both $\theta$ and $\phi$ could be performed jointly, we decide to alternate both steps. To initialize the process, we arbitrarily fit $h_k^{\phi}$ for $E$ epochs on the raw outputs, yielding
\begin{align}
\label{eq:initial_prior_fit}
    \phi^{(0)} &= \arg\min_{\phi} \mathcal{L}_{\textit{prior}}(h_k^{\phi}, f; \textit{LS}),
\end{align}
where $\mathcal{L}_{\textit{prior}}(h_k^{\phi}, f; \textit{LS}) = \frac{1}{N}\sum_{i=1}^{N}(h_k^{\phi}(\x_k^{(i)}) - y^{(i)})^2$. Then, we fit the ML model under the constraint that its PD should be close to the current model $h_k^{\phi^{(0)}}$. Every $S \in \N^+$ epochs, we update $\phi$ to minimize \eqref{eq:sample_pd_mse}, for the current fit of $\theta$. We further repeat the previous two steps. 

It is important to note that, given the definition of \eqref{eq:sample_pd_mse}, we can compute $\mathcal{L}_{\textit{PD}}$ using a dataset $\textit{LS}_{\textit{PD}} = \{\x_k^{(j)}\}_{j=1}^{N_{\textit{PD}}}$ different from $\textit{LS}$:
\begin{align*}
     \mathcal{L}_{\textit{PD}}(h^{\theta}, h_k^{\phi}; \textit{LS},\textit{LS}_{\textit{PD}})
    =  
    \frac{1}{N_{\textit{PD}}}\sum_{j=1}^{N_{\textit{PD}}}(\textit{PD}_N(h^{\theta}, \x_k^{(j)})-h_k^{\phi}(\x_k^{(j)}))^2.
\end{align*}
While $\textit{LS}$ is used to compute the value of $\textit{PD}_N$ at a given point (i.e. to obtain samples from $p(\x_{-k})$), $\textit{LS}_{\textit{PD}}$ is a dataset of said given points $\x_k$. The utility of constructing separate datasets for these two steps is further explained in Section \ref{subsec:constraint_configurations}. Algorithm \ref{alg:pd_driven_training} summarizes the training procedure, where steps of gradient descent $\nabla_{\cdot}$ are performed on mini-batches.

\begin{algorithm}[t]
    \caption{Partial Dependence Steering}
    \label{alg:pd_driven_training}
    \begin{algorithmic}
        \State {\bfseries Input:} $\textit{LS} = \left\{(\x^{(i)}, y^{(i)})\right\}_{i=1}^N, LS_{\textit{PD}} = \left\{\x_k^{(j)}\right\}_{j=1}^{N_{\textit{PD}}}$
        \State {\bfseries Parameters:} $E > S \geq 0, \tau_1 > 0, \tau_2 > 0, \lambda > 0$
        \State
        \For{$e=1$ {\textbf{to}} $E$}
        \State $\phi^{(e)} \leftarrow \phi^{(e-1)} - \tau_1 \nabla_{\phi}\mathcal{L}_{\textit{prior}}(h_k^{\phi^{(e-1)}}, f; \textit{LS})$
        \EndFor
        \State $\phi^{(0)} \leftarrow \arg\min_{e}\mathcal{L}_{\textit{prior}}(h_k^{\phi^{(e)}}, f; \textit{LS})$
        \State
        \State $s = 0, r = 1$
        \For{$e=1$ {\bfseries to} $E$}
        \If{$s < S$}
        \State \!\!\!\!$\theta^{(e)} \!\leftarrow\! \theta^{(e-1)} \!-\! \tau_2 \nabla_{\theta}\mathcal{L}_{\lambda}(h^{\theta^{(e-1)}}\!, h_k^{\phi^{(r-1)}}\!, f; \textit{LS}, \textit{LS}_{\textit{PD}})$
        \State \!\!\!\!$s = s + 1$
        \Else
        \State \!\!\!\!$\phi^{(r)} \!\leftarrow \phi^{(r-1)} \!- \tau_2 \nabla_{\phi}\mathcal{L}_{\textit{PD}}(h^{\theta^{(e-1)}}\!, h_k^{\phi^{(r-1)}}; \textit{LS}, \textit{LS}_{\textit{PD}})$ \\
        \State \!\!\!\!$\theta^{(e)} \leftarrow \theta^{(e-1)}$
        \State \!\!\!\!$s = 0, r = r + 1$
        \EndIf
        \EndFor
        \State $\theta \leftarrow \arg\min_{e}\mathcal{L}_{\textit{tot}}(h^{\theta^{(e)}}, f; \textit{LS})$
        \State \textbf{return} $\theta$
    \end{algorithmic}
\end{algorithm}


\subsection{Constraint configurations}
\label{subsec:constraint_configurations}
We consider different scenarios, where we vary both the knowledge level and $\textit{LS}_{\textit{PD}}$. In all cases, the shape of $\textit{PD}_{\infty}(f, \x_k)$ is assumed to be known through $h_k^{\phi}(\x_k)$. However, the \textit{values} of the parameters $\phi$ can either be known exactly (i.e., $\phi =\phi^*$, denoted \textit{ground truth prior}), or have to be learned (i.e., $\phi = \hat{\phi}$, denoted \textit{approximated prior}), as in Algorithm \ref{alg:pd_driven_training}. Furthermore, we can define $\textit{LS}_{\textit{PD}}$ in two ways: $\textit{LS}_{\textit{PD}} = \{\x_k^{(i)} \mid \x^{(i)} \in \textit{LS}\}_{i=1}^{N}$ (\textit{in-distribution constraint}), where the PD of the model will be constrained at positions defined by the learning sample, i.e. $\x_k \sim p(\x_k)$, or as a set of evenly spaced values between lower and upper bounds $l, u \in \R^{d_{\x_k}}$, i.e. $\textit{LS}_{\textit{PD}} = \{\x_k^{(j)} \in \R^{d_{\x_k}} \mid \x_k^{(j)} = l + (j-1) \frac{u-l}{N_{\textit{PD}}-1} \}_{j=1}^{N_{\textit{PD}}}$ (\textit{linspace constraint}). Such bounds can be the minimum and maximum observed values for $\x_k$ in the dataset, or derived from prior knowledge. Constructing a separate dataset $\textit{LS}_{\textit{PD}}$ allows to constrain on regions that may be poorly sampled, thereby helping the model to generalize more effectively. In particular, this could help in a scenario where the model is later on fed with out-of-distribution data.

\section{Experiments}
\label{sec:experiments}
In this section, we study the performance of our PD-based steering method, in terms of both predictive performance and quality of the resulting PD, through test set estimates of \eqref{eq:sample_mse_def} and \eqref{eq:sample_pd_mse}, with $\phi = \phi^*$, which we refer to as $\mathcal{L}_{\textit{tot}}$ and $\mathcal{L}_{\textit{PD}}$. We compare our four settings to unconstrained training on synthetic and real regression problems, as well as on a problem of dynamical systems forecasting. We study the effect of training sample size on both measures, and analyze the impact of shifting the domain of $\x_k$ at test time. We also compare the performance of our method against that of physics-informed ML models, and eventually study the behavior of models constrained with mis-specified prior knowledge.

\paragraph{Dataset and model hyperparameters}
For synthetic regression problems, we generate training and validation sets of successively 300, 150 and 50 samples each. All methods are evaluated on a fixed test set of 10\,000 samples. For the lower and upper bounds of linspace constraints, we use the bounds of the input domain, and we arbitrarily fix the resolution to 25 for PD functions of two variables (producing a $25 \times 25$ grid) and to 256 for single-variable PD functions. For both constraining types, we draw random mini-batches from $\textit{LS}_{\textit{PD}}$ at each epoch. We consider the following settings (see Section \ref{subsec:constraint_configurations}):
\begin{itemize}
    \item Linspace constraint on the ground truth prior, denoted as $\phi = \phi^*$ (L),
    \item In-distribution constraint on the ground truth prior, denoted as $\phi = \phi^*$ (ID),
    \item Linspace constraint on the approximated prior, denoted as $\phi = \hat{\phi}$ (L),
    \item In-distribution constraint on the approximated prior, denoted as $\phi = \hat{\phi}$ (ID).
\end{itemize}

We fix the architecture of the neural network to three layers of increasing hidden size (32, 64 and 128), with a fixed training batch size of 16 samples. We optimize the learning rate $\tau_2$ in $\{\num{5e-2}, \num{5e-3}, \num{5e-4}\}$ and the coefficient $\lambda$ in $\{0.25, 0.5, 0.75\}$, selecting the values that minimize $\mathcal{L}_{\textit{tot}}$ on the validation set. We set the number of epochs $E = 1000$, $\tau_1 = 1$, $S = 50$.

\paragraph{Test-time domain shift}
To implement this behavior, in synthetic problems, we first define an interpolation range on a particular variable in $\x_k$. We then generate training and validation sets by sampling from the original distribution $p(\x)$, rejecting any sample whose value for the selected variable lies outside the aforementioned range. The test set is identical for interpolation and extrapolation, such that it contains samples with values both inside and outside the interpolation range.

\subsection{Friedman problem}
We consider the following problem:
\begin{align*}
    y = \psi_{0} \sin(\psi_{1} x_0 x_1) + \psi_{2} (x_2 - \psi_{3})^2 + \psi_{4} x_3 + \psi_{5} x_4 + \sum_{j=5}^{9}0 x_j + \varepsilon,
\end{align*}
where $x_j \sim \mathcal{U}(0, 1), j=0, \dots 9$, $\psi = [10, \pi, 20, 0.5, 10, 5]$ and $\varepsilon \sim \mathcal{N}(0, 1)$ \citep{friedman1983multidimensional}. In what follows, we consider constraining the PD of $h^{\theta}$ w.r.t. $x_0, x_1$ to the following function:
\begin{align*}
    h_k^{\phi}(x_0, x_1) &= \; 
        \phi_{0}[\phi_2 \sin(\phi_{3} x_0 x_1)] + \phi_1,
\end{align*}
with $\phi^* = [1, C_1, \psi_0, \psi_1]$ a potential ground truth parameter vector, where:
\begin{align*}
    C_1 &= \mathbb{E}_{\x_{-k}}\left[\psi_{2} (x_2 - \psi_{3})^2 + \psi_{4} x_3 + \psi_{5} x_4\right].
\end{align*}
We derive similar discussions for constraining $\textit{PD}_N(h^{\theta}, x_2)$ and $\textit{PD}_N (h^{\theta}, x_3, x_4)$ in Appendix~\ref{app:friedman}.

\begin{figure}[ht]
    \centering
    \includegraphics[width=\linewidth]{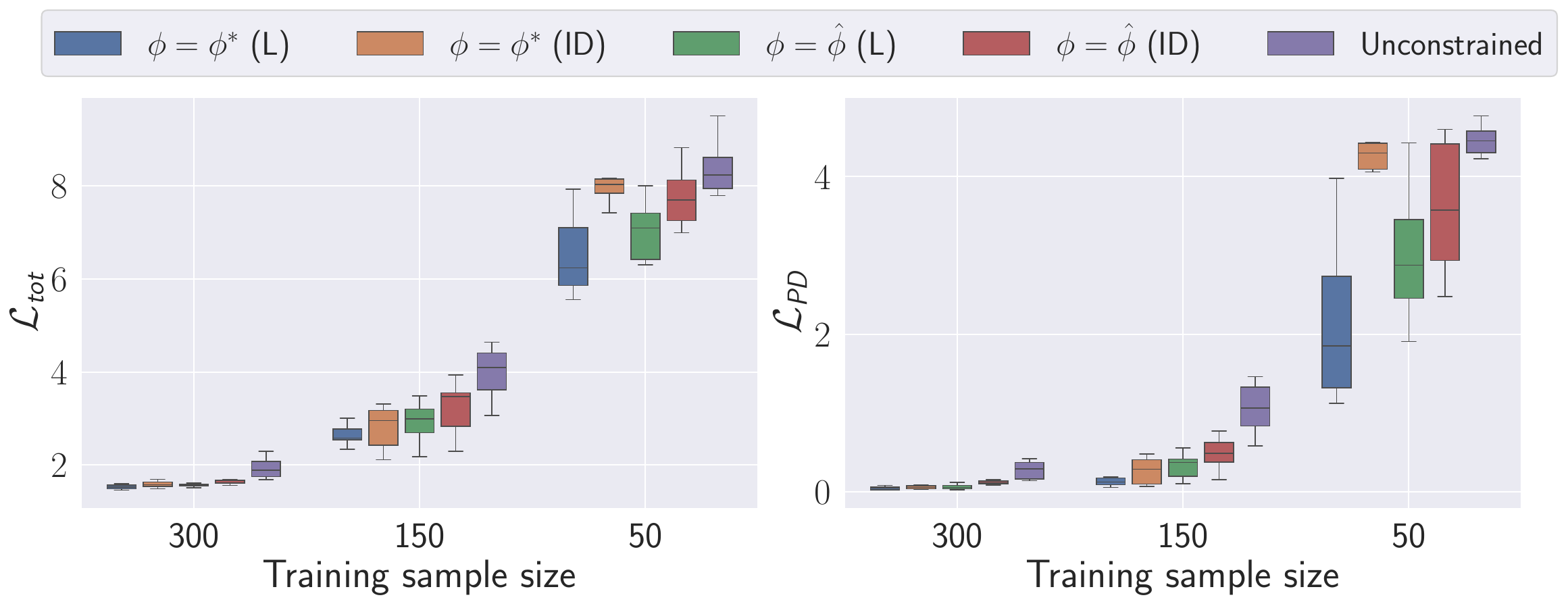}
    \caption{Evolution of $\mathcal{L}_{\textit{tot}}$ (left) and $\mathcal{L}_{\textit{PD}}$ (right) w.r.t. the training sample size, on the Friedman problem, constraining $\textit{PD}_N(h^{\theta}, x_0, x_1)$. Each boxplot summarizes the results on the test set, over 10 different training initializations.}
    \label{fig:friedman_first_mse_pd_mse}
\end{figure}

\paragraph{Sample size effect}
As a general trend, we observe in Figure \ref{fig:friedman_first_mse_pd_mse} that decreasing the learning sample size degrades both measures of performance, for all methods. Furthermore, constraining the PD of the model improves both measures compared to unconstrained training, for all training sizes.

\begin{figure}[ht]
    \centering
    \includegraphics[width=\linewidth]{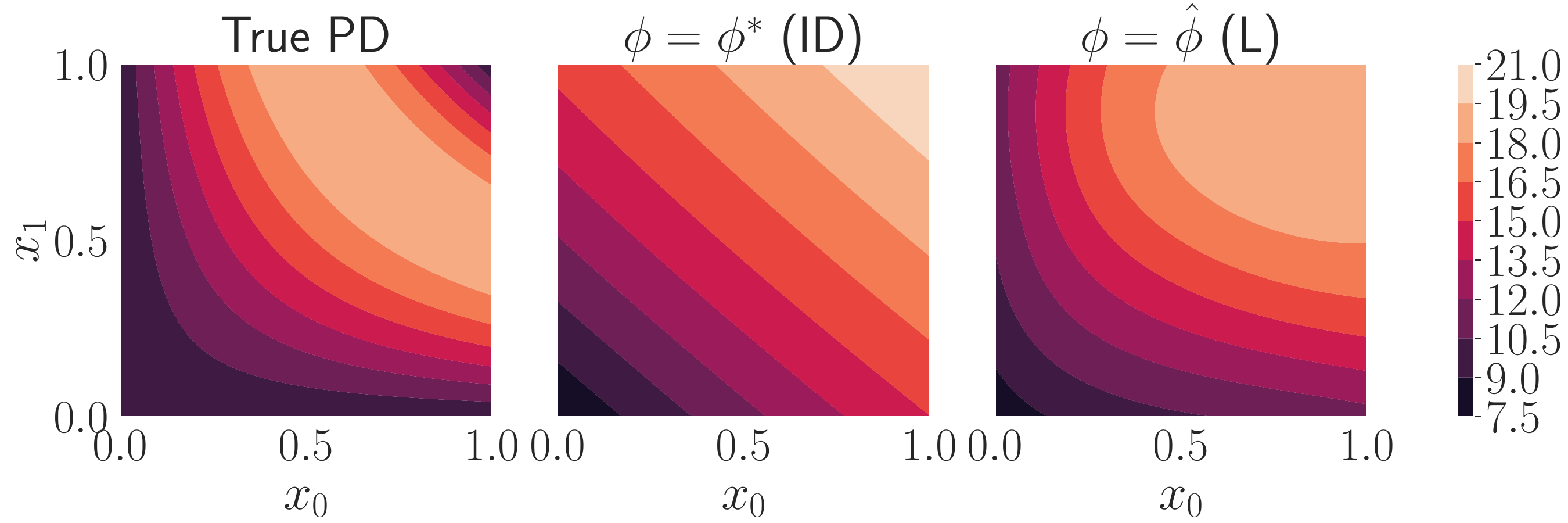}
    \caption{Examples of PD estimation for $\textit{PD}_N(h^{\theta}, x_0, x_1)$. The figure displays the true PD (left), the PD after in-distribution constraining on the ground truth PD (middle), and the PD after linspace constraining on the approximated prior (right). Both have been trained with 50 samples, in the interpolation setting.}
    \label{fig:friedman_pd_first_50_samples_interpolation}
\end{figure}

As could be expected, linspace constraining on the true PD function (blue boxes) achieves the best results for both metrics, whatever the sample size. Nevertheless, it is interesting to note that linspace constraining with an approximated prior $\phi = \hat{\phi}$ (green boxes) performs on par with in-distribution constraining on true PD samples (orange boxes), and even outperforms the latter when data becomes very scarce (50 samples), which highlights the benefit of constraining the PD in poorly-sampled regions of the input space. This can be verified by looking at Figure \ref{fig:friedman_pd_first_50_samples_interpolation}, which highlights the difference in the PD estimates.

\paragraph{Test-time domain shift}
We further test the robustness of constrained models when observing test-time domain shift in $\x_0$: samples for which $\x_0 > 0.75$ are rejected from the training and validation sets.

\begin{figure}[ht]
    \centering
    \includegraphics[width=\linewidth]{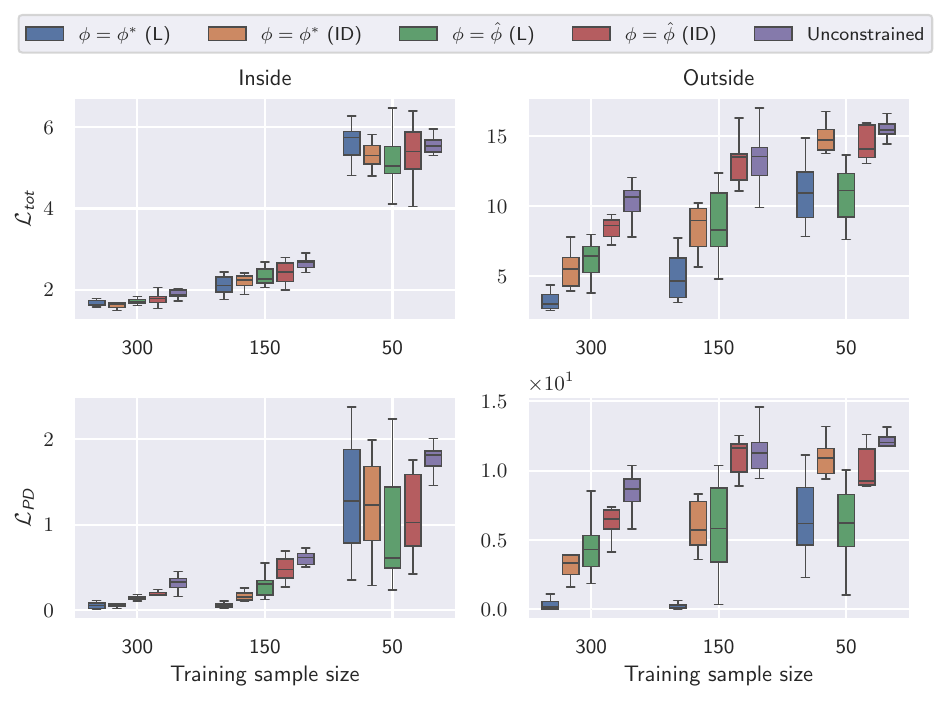}
    \caption{Evolution of $\mathcal{L}_{\textit{tot}}$ (top plots) and $\mathcal{L}_{\textit{PD}}$ (bottom plots) w.r.t. the training sample size, on the Friedman problem, constraining $\textit{PD}_N(h^{\theta}, x_0, x_1)$. \textit{Inside} metrics are computed on samples for which $x_0 \leq 0.75$ (left plots), while \textit{Outside} metrics are computed on samples where $x_0 > 0.75$ (right plots). Each boxplot summarizes the results on the test set, over 10 different training initializations.}
    \label{fig:friedman_first_inside_outside}
\end{figure}

We can observe in Figure \ref{fig:friedman_first_inside_outside} that the relative ranking of training methods inside the interpolation range is similar to that observed in Figure \ref{fig:friedman_first_mse_pd_mse}. Note that this similarity is not necessarily expected as training settings are quite different from the previous experiment. Indeed, for in-distribution constraints, given that the PD of the model is now constrained only up to $x_0 = 0.75$, the model can produce different results from the one constrained on the entire domain, even inside the interpolation range. Looking at $\mathcal{L}_{\textit{tot}}$ inside the interpolation range, we can notice that we now obtain overall lower errors compared to the previous setting, whatever the training scheme. This is expected since training methods now have access to more data samples in a narrower space, but it could also mean that the output signal for $x_0 > 0.75$ is more complex to capture. For both metrics, it is clear that PD steering helps to dampen the impact of moving outside the interpolation range, once again much more when constraining on evenly-spaced points.

\subsection{Product problem}
We consider the following problem:
\begin{align*}
    y = \psi_0 x_0^3 \cos(\psi_1 x_0) \frac{(x_1 - 1)^2}{2} + \sum_{j=2}^{9}0 x_j + \varepsilon,
\end{align*}
where features are correlated with zero mean, variances and covariances equal to $0.5$ and $0.25$, respectively, $\psi = [15, \pi]$, and $\varepsilon \sim \mathcal{N}(0, 1)$, with the following PD function:
\begin{equation*}
    h_k^{\phi}(x_0) = \phi_0 [\phi_2 x_0^3 \cos(\phi_3 x_0)] + \phi_1.
\end{equation*}
A potential ground truth parameter vector is $\phi^* = [C, 0, \psi_0, \psi_1]$, where:
\begin{equation*}
    C = \mathbb{E}_{x_1}\left[\frac{(x_1 - 1)^2}{2}\right].
\end{equation*}

\begin{figure}[ht]
    \centering
    \includegraphics[width=\linewidth]{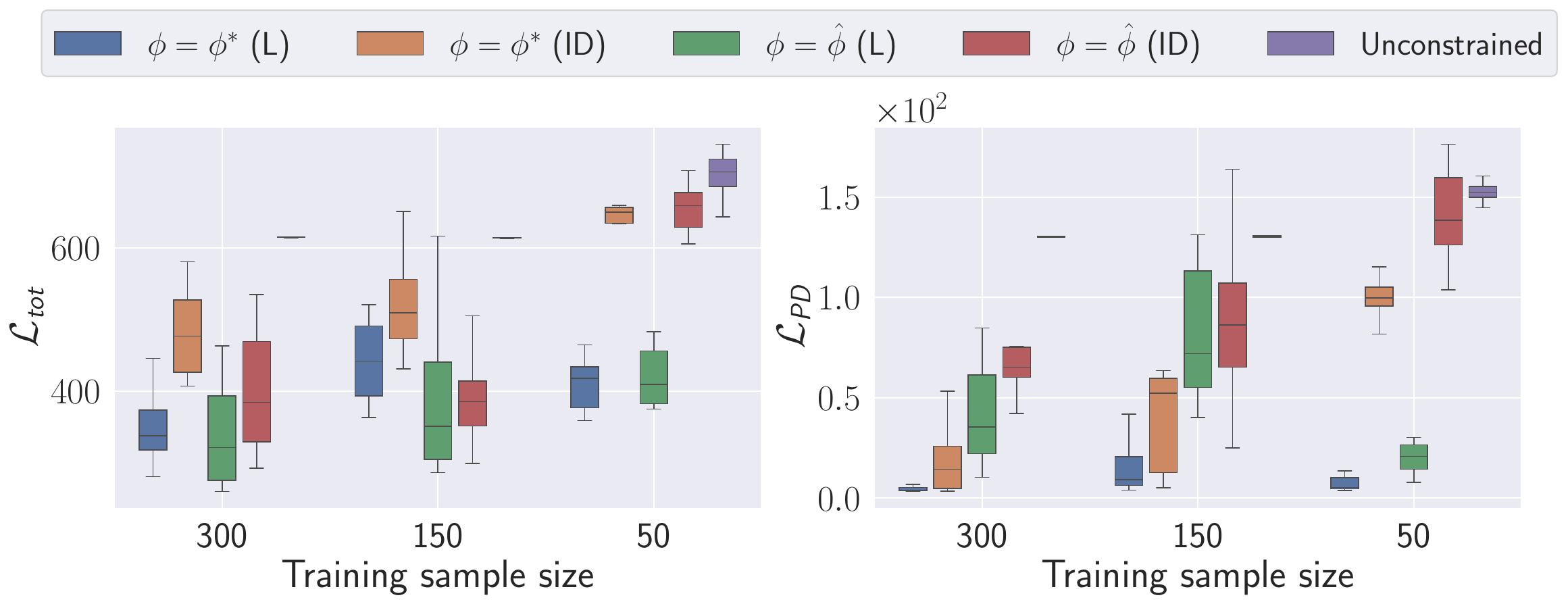}
    \caption{Evolution of $\mathcal{L}_{\textit{tot}}$ (left) and $\mathcal{L}_{\textit{PD}}$ (right) w.r.t. the training sample size, on the product problem, constraining $\textit{PD}_N(h^{\theta}, x_0)$. Each boxplot summarizes the results on the test set, over 10 different training initializations.}
    \label{fig:product_mse_pd_mse}
\end{figure}

\paragraph{Sample size effect}
As was the case for the Friedman problem, we can observe in Figure \ref{fig:product_mse_pd_mse} that model performance degrades as the training size decreases, although more slightly. We notice that our regularization methods vastly help to learn more robust models with limited amounts of data compared to unconstrained training. Indeed, the test mean-squared errors of in-distribution constrained models (orange and red boxes) on 50 samples are roughly at the same level as those of the unconstrained model trained on 300 samples, and linspace constrained models (blue and green boxes) vastly outperform the latter. We also notice that the PD obtained for most constrained models on 50 samples is even better than that of the unconstrained neural network on 300 samples. Figure \ref{fig:product_mse_pd_mse} also highlights the benefits of constraining on a set of evenly-spaced points, as a wider region of the input domain is covered, even against in-distribution constraining on the ground truth PD function (orange boxes). Interestingly, linspace constraining on an approximated prior (green boxes) can produce more accurate predictions compared to linspace constraining on the true function (blue boxes). This can be explained by the fact that, by approximating the PD function, the former allows for more flexibility to reduce the mean-squared error on the outputs, at the cost of a slightly worse PD estimation.

\begin{figure}[ht]
    \centering
    \includegraphics[width=\linewidth]{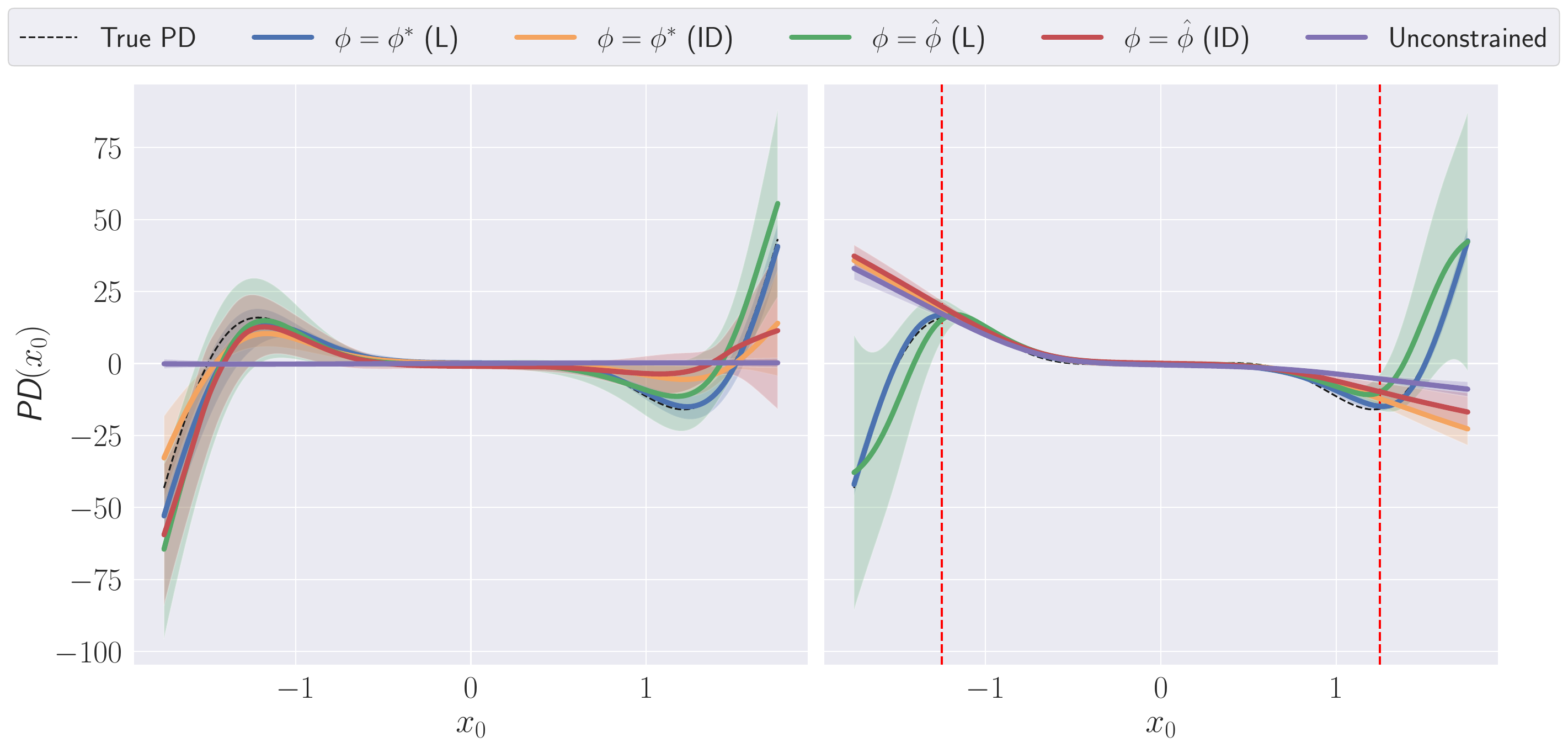}
    \caption{The figure displays the mean PD estimations for $\textit{PD}_N(h^{\theta}, x_0)$ obtained with each method, for the product problem, in the interpolation (left) and extrapolation settings (right). All have been trained on 150 samples. Mean and standard deviations are computed over 10 different training initializations. The dotted red lines represent the boundaries of the interpolation range.}
    \label{fig:product_150_samples_pd_comparison}
\end{figure}

We represent in Figure \ref{fig:product_150_samples_pd_comparison} (left) the different PD estimations obtained on average with each training method, with a training set of 150 samples. We notice that each constrained model approximates the true PD function quite well, while the unconstrained neural network has more difficulties in approximating the boundary regions, where the PD function varies more sharply.

Even though PD plots (PDP) as interpretation tools are expected to deteriorate and be less faithful in the presence of (strongly) correlated features, due to the averaging of predictions over possibly unrealistic input samples \citep{molnar2020interpretable}, we observe that this does not impact our method as we constrain models with the true algebraic form of the PD.

\begin{figure}[ht]
    \centering
    \includegraphics[width=\linewidth]{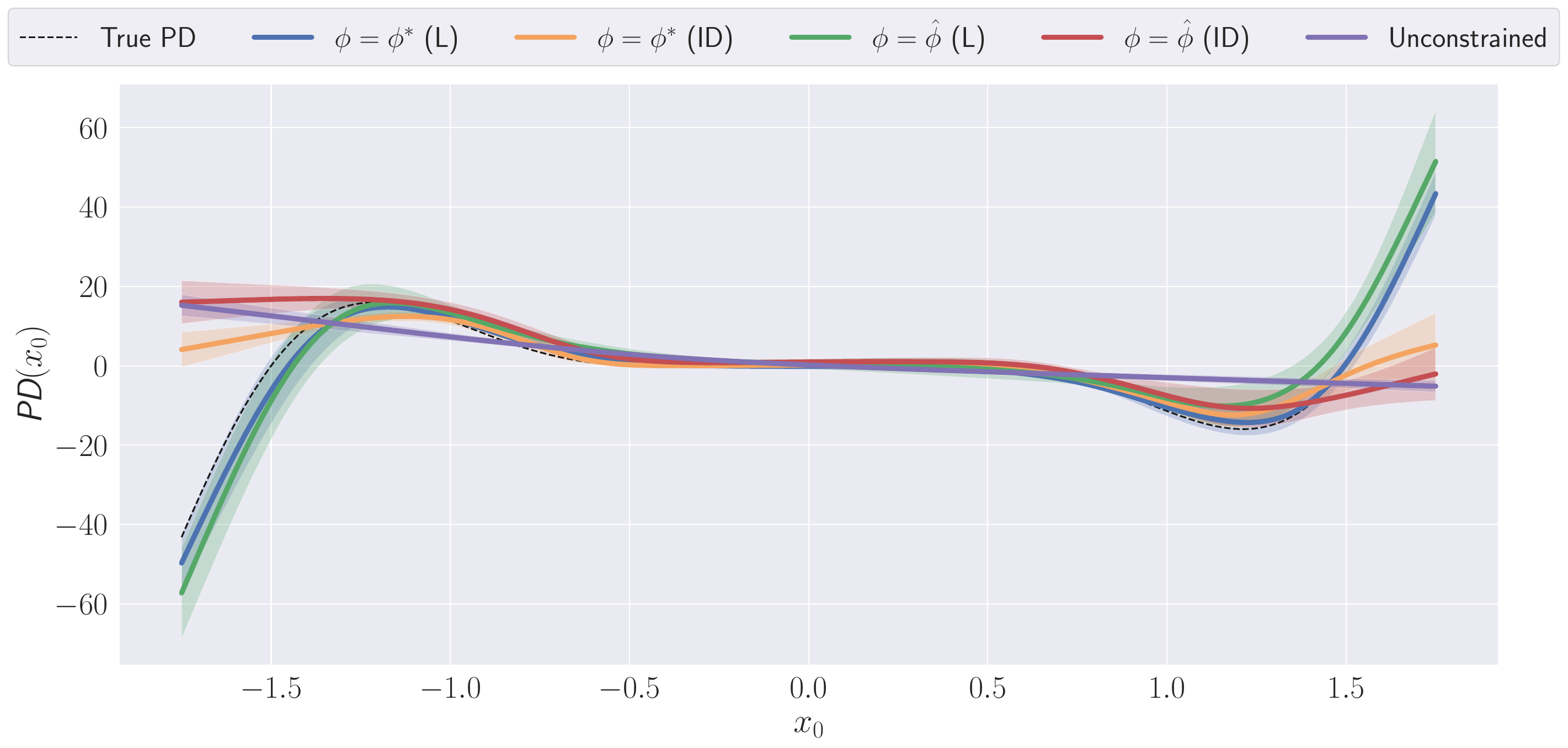}
    \caption{The figure displays the mean PD estimations for $\textit{PD}_N(h^{\theta}, x_0)$ obtained with each method, for the product problem, in the interpolation setting. All methods have been trained on 50 samples. Mean and standard deviations are computed over 10 different training initializations.}
    \label{fig:50_interpolation_product_true_vs_est_pd}
\end{figure}

In Figure \ref{fig:50_interpolation_product_true_vs_est_pd}, we show that in-distribution constraining fails to provide an accurate PD estimate, compared to linspace constraining, when the number of training points is scarcer ($N=50$).

\paragraph{Test-time domain shift}
We study the impact of shifting the domain of $x_0$ at test time: samples for which $x_0 \notin [-1.25, 1.25]$ are rejected from the training and validation sets, which roughly corresponds to the switch of concavity of the PD function, on both sides.

\begin{figure}[ht]
    \centering
    \includegraphics[width=\linewidth]{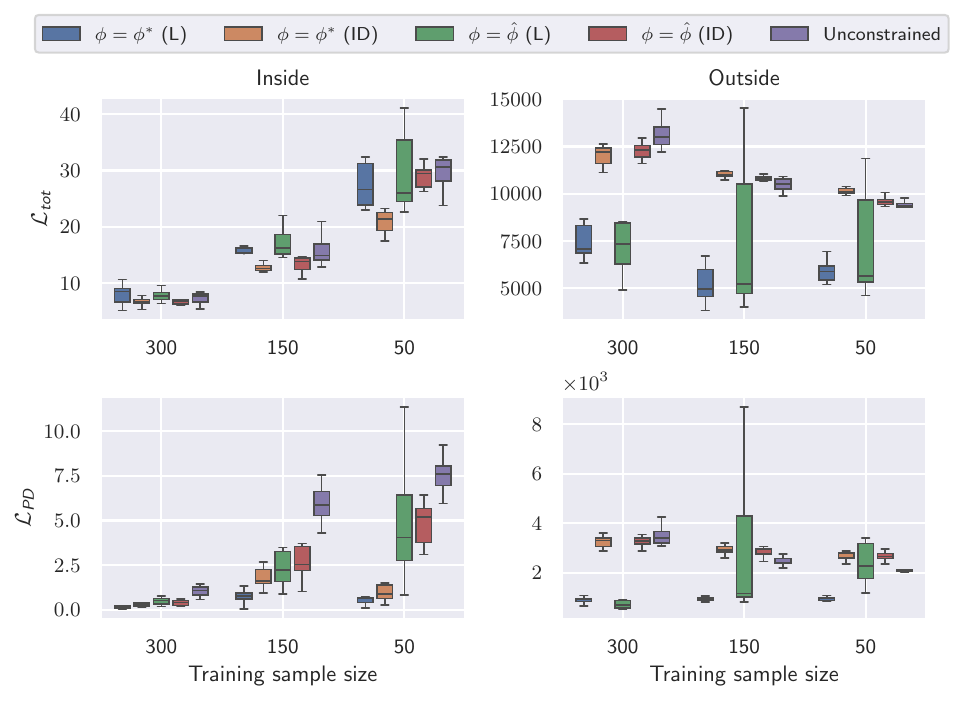}
    \caption{Evolution of $\mathcal{L}_{\textit{tot}}$ (top plots) and $\mathcal{L}_{\textit{PD}}$ (bottom plots) w.r.t. the training sample size, on the product problem, constraining $\textit{PD}_N(h^{\theta}, x_0)$. \textit{Inside} metrics are computed on samples for which $x_0 \in [-1.25, 1.25]$ (left plots), while \textit{Outside} metrics are computed on samples outside that range (right plots). Each boxplot summarizes the results on the test set, over 10 different training initializations.}
    \label{fig:larger_product_first_inside_outside}
\end{figure}

In Figure \ref{fig:larger_product_first_inside_outside}, we can observe that for training sets of 300 and 150 samples, unconstrained training performs on par (if not sometimes better) with constrained methods, when samples lie inside the interpolation range. By comparing with the mean-squared errors reported in Figure \ref{fig:product_mse_pd_mse}, we hypothesize that the output region in the center of the domain is easier to approximate than the borders, which is also suggested by the true PD function shape in Figure \ref{fig:product_150_samples_pd_comparison}. This claim can be further verified by looking at both metrics outside the interpolation range. Indeed, linspace-constrained models (blue and green boxes) vastly outperform unconstrained ones in both measures, producing similar results whatever the training set size, although with some variance when the PD parameters need to be estimated. On the other hand, it is interesting to note that in-distribution constraining (orange and red boxes) only slightly improves performance (or even worsens) against unconstrained training. This can be again explained by the fact that the PD of the model is constrained only inside the interpolation domain, while most modeling complexity lies at the boundaries of the domain, which is what we verify in Figure \ref{fig:product_150_samples_pd_comparison} (right). Indeed, the PD estimations of in-distribution training are actually very similar to those obtained after unconstrained training.

\subsection{Concrete compressive strength prediction}
This real-world dataset collects 1\,030 samples relating eight characteristics of concrete components with the resulting compressive strength \citep{concrete_compressive_strength_165}. A complete description of all the features can be found in Appendix \ref{app:concrete_variables_description}.

Based on the dataset introductory paper \citep{yeh1998modeling}, we assume the relationship between age and compressive strength to be logarithmic, i.e.,
\begin{align*}
    h_k^{\phi}(\text{age}) = \phi_0\ln(\text{age}) + \phi_1.
\end{align*}

Given that this experiment is based on real data, we do not have access to ground truth values of $\phi$, therefore we can only experiment with the approximate constraining methods and cannot measure $\mathcal{L}_{\textit{PD}}$. We choose to use 100, 50 and 25 training (and validation) samples, leaving the rest as test set, and set the batch size to 5 samples. For each training size, we generate 20 different dataset splits.

\begin{figure}[ht]
    \centering
    \includegraphics[width=\linewidth]{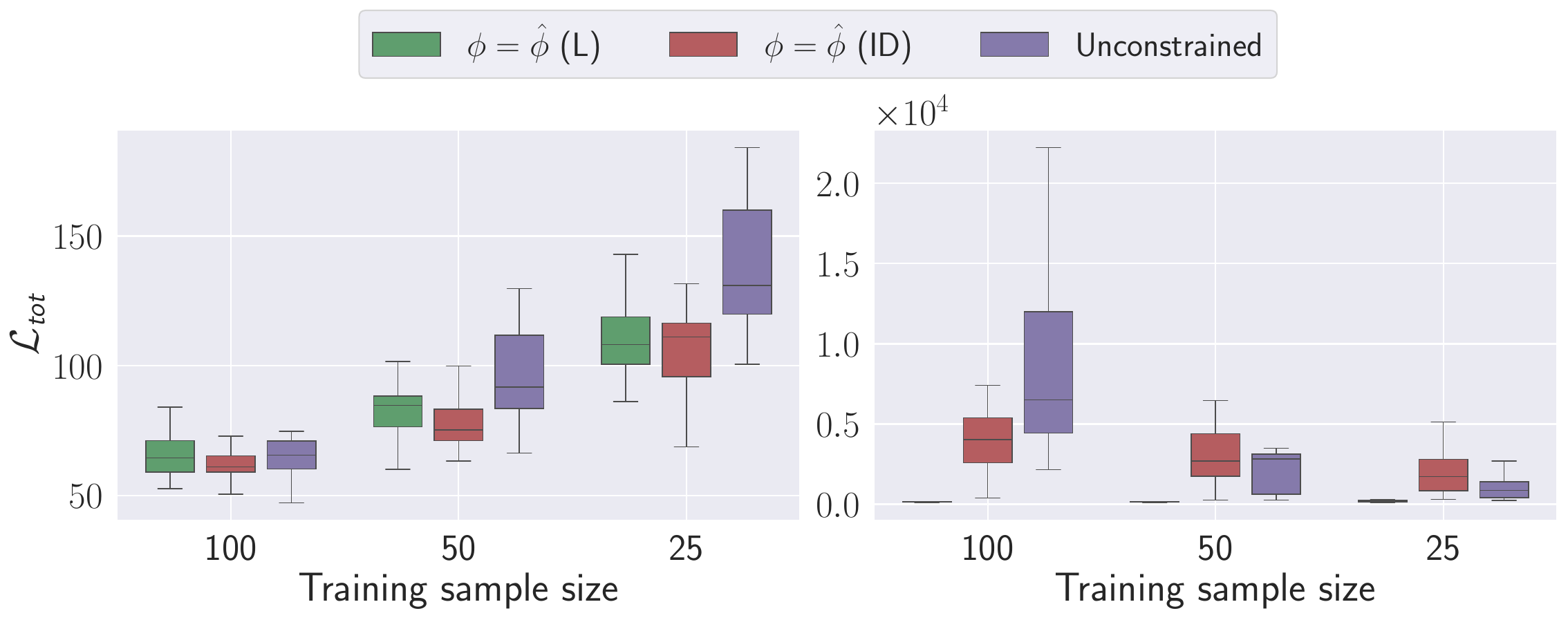}
    \caption{Evolution of $\mathcal{L}_{\textit{tot}}$ w.r.t. the training sample size, on the concrete problem, constraining $\textit{PD}_N(h^{\theta}, \text{age})$, in the interpolation setting (left) and extrapolation setting (right). Each boxplot summarizes the results on a test set, over 20 different dataset splits.}
    \label{fig:concrete_mse_interpolation_extrapolation}
\end{figure}

\paragraph{Sample size effect}
We observe in Figure \ref{fig:concrete_mse_interpolation_extrapolation} (left) that constraining the PD of the network to be logarithmic helps to reduce the errors made in the final predictions, even more for smaller training sizes. With enough training samples, the gap w.r.t. unconstrained training is closed.

\paragraph{Test-time domain shift}
In this setting, we have removed from training and validation sets samples whose age value was larger than 15 days, which allows to have a shared test set of 706 samples for all extrapolation experiments.

We observe in Figure \ref{fig:concrete_mse_interpolation_extrapolation} (right) that unconstrained models suffer from shifting outside the age range considered during training. With enough training samples, in-distribution constraining is able to dampen this effect, but it then  levels with unconstrained training, and even worsens for smaller training sizes. Linspace constraining, however, vastly outperforms both in-distribution constraining and unconstrained training, for all training sizes, and reaches performance close to that obtained in the interpolation setting.

\begin{figure}[ht]
    \centering
    \includegraphics[width=\linewidth]{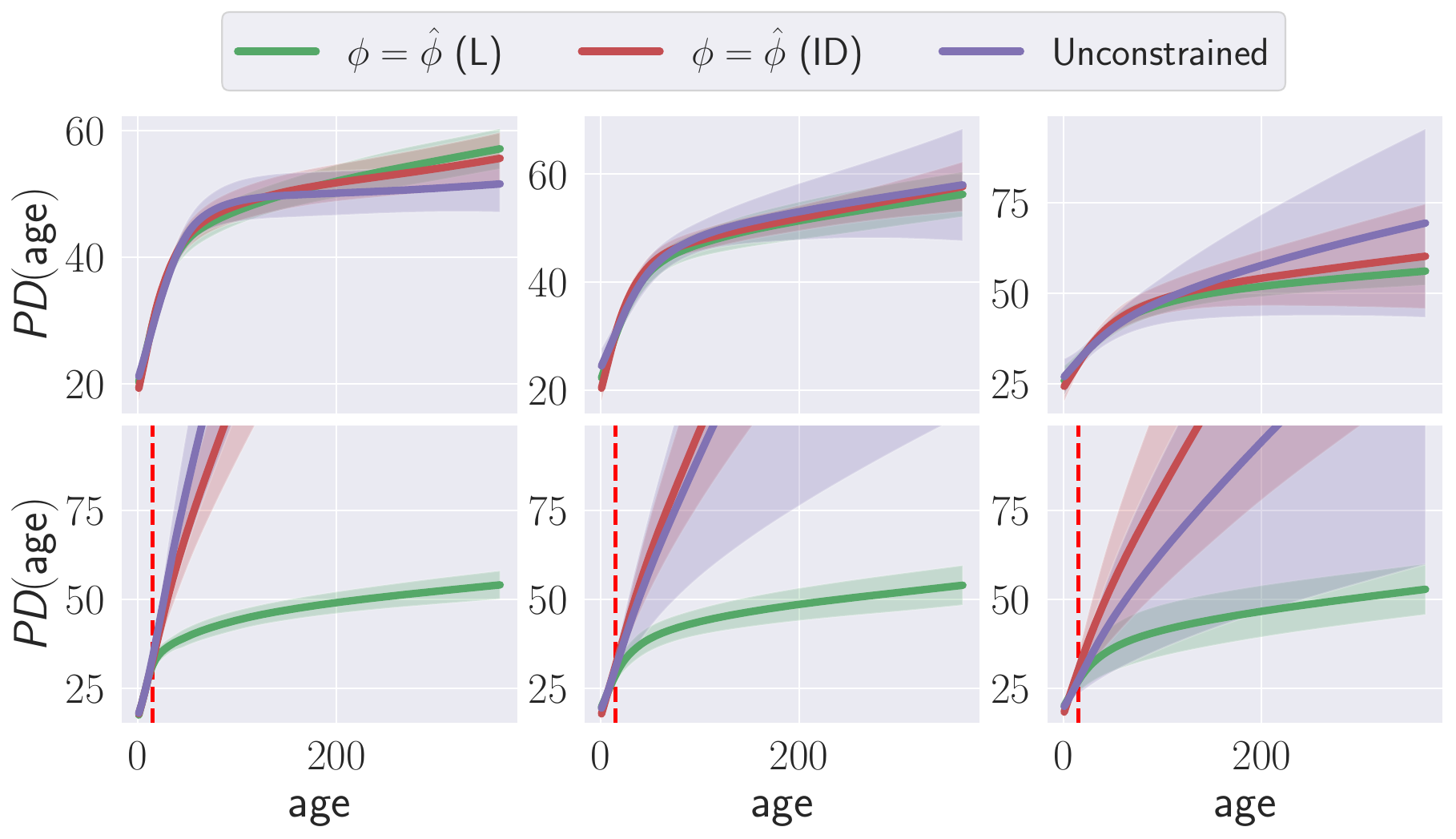}
    \caption{The figure displays the mean PD estimations for $\textit{PD}_N(h^{\theta}, \text{age})$ obtained with each method, for the compressive strength problem, in the interpolation (top) and extrapolation settings (bottom). All methods have been trained respectively on 100 samples (left), 50 samples (middle), 25 samples (right). Mean and standard deviations are computed over 20 different dataset splits. The dotted red lines represent the boundaries of the interpolation range.}
    \label{fig:concrete_all_sizes_pd_comparison}
\end{figure}

In Figure \ref{fig:concrete_all_sizes_pd_comparison}, we provide the mean PD estimations for all methods, both in the interpolation (top) and extrapolation (bottom) settings. In the interpolation setting, we can observe that, with enough training data (100 samples), the PD estimations of all models follow the same logarithmic trend. However, when reducing the number of training samples, we notice that those of unconstrained models tend to diverge and show increased variance.
In the extrapolation setting, the situation is radically different. Indeed, both unconstrained models and in-distribution constraining fail to capture the logarithmic shape outside the training domain, which confirms the poor performance that was observed for such methods in Figure \ref{fig:concrete_mse_interpolation_extrapolation} (right). For this problem, it is clear that the linspace constraint helps to better model the shape of the desired PD, thereby reducing the prediction error, as was observed in Figure \ref{fig:concrete_mse_interpolation_extrapolation} (right).

\subsection{PHALK dataset}
This dataset relates environmental basin characteristics, alkalinity and pH values from a very large number of inland water sites \citep{phalk}, for a total of 1\,017\,498 samples. A complete description of all the features can be found in Appendix \ref{app:phalk_variables_description}.

As mentioned in the introductory paper \citep{batalla2026global}, we assumed a logarithmic relationship between pH and alkalinity. Therefore, we decided to train a model to predict pH from the other features, and constrained its partial dependence w.r.t. alkalinity to be logarithmic, i.e.,
\begin{align*}
    h_k^{\phi}(\text{alkalinity}) = \phi_0\ln(\text{alkalinity}) + \phi_1.
\end{align*}
As for the previous experiment, we use 100, 50 and 25 training (and validation) samples, leaving the rest as test set, and set the batch size to 5 samples. For each training size, we generate 20 different dataset splits.

\begin{figure}[ht]
    \centering
    \includegraphics[width=\linewidth]{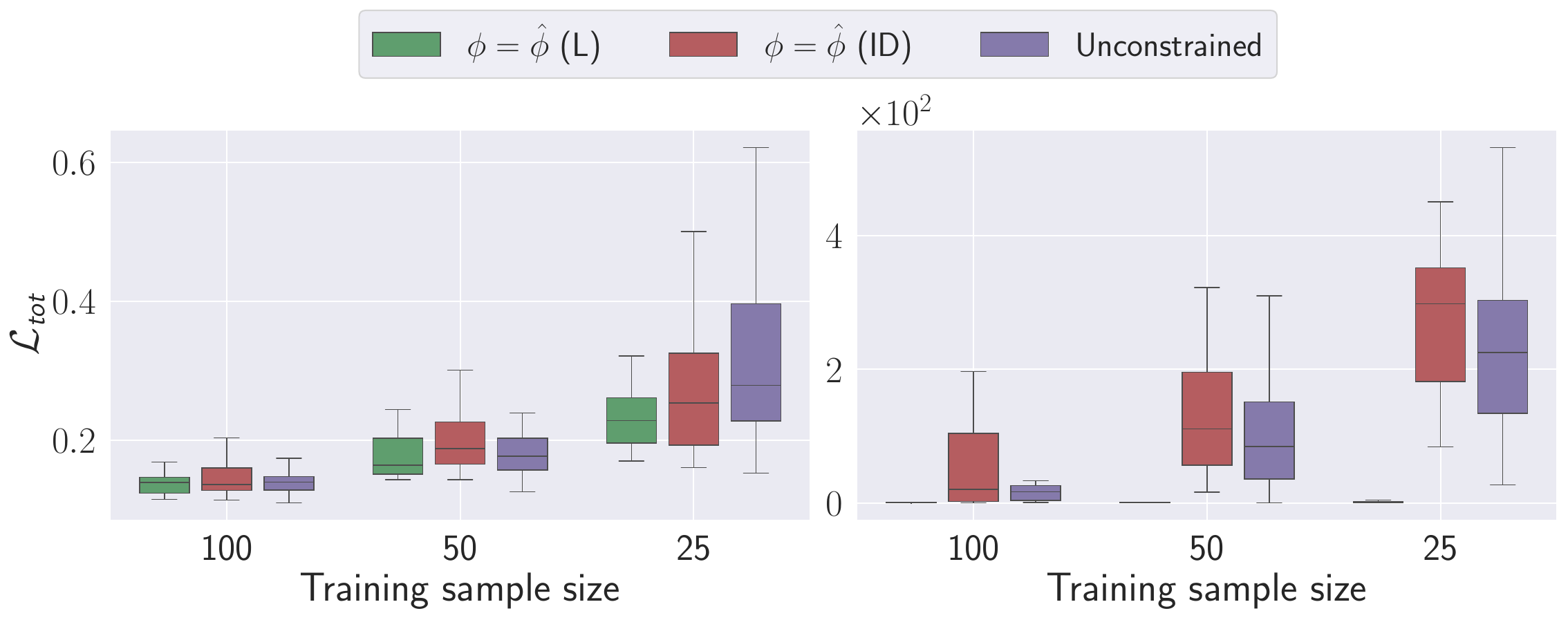}
    \caption{Evolution of $\mathcal{L}_{\textit{tot}}$ w.r.t. the training sample size, on the PHALK dataset, constraining $\textit{PD}_N(h^{\theta}, \text{alkalinity})$, in the interpolation setting (left) and extrapolation setting (right). Each boxplot summarizes the results on a test set, over 20 different dataset splits.}
    \label{fig:phalk_mse_interpolation_extrapolation}
\end{figure}

\paragraph{Sample size effect}
Similarly to the concrete compressive strength prediction problem, we observe in Figure \ref{fig:phalk_mse_interpolation_extrapolation} (left) that constraining the PD of the model to be logarithmic improves final performance, especially for smaller training sizes.

\paragraph{Test-time domain shift}
As typical alkalinity values observed in the dataset lie in $[0, 8000]$, we set a threshold to 250 for the extrapolation setting, i.e., training (and validation) samples only have alkalinity values lower than 250, while test samples have alkalinity values higher than 250.

We observe in Figure \ref{fig:phalk_mse_interpolation_extrapolation} (right) that, once again, linspace constraining vastly outperforms both in-distribution constraining and unconstrained training. For this problem, in-distribution constraining yields the worst models, whereas linspace constraining seems to reach performance levels close to those observed in the interpolation setting.

\begin{figure}[ht]
    \centering
    \includegraphics[width=\linewidth]{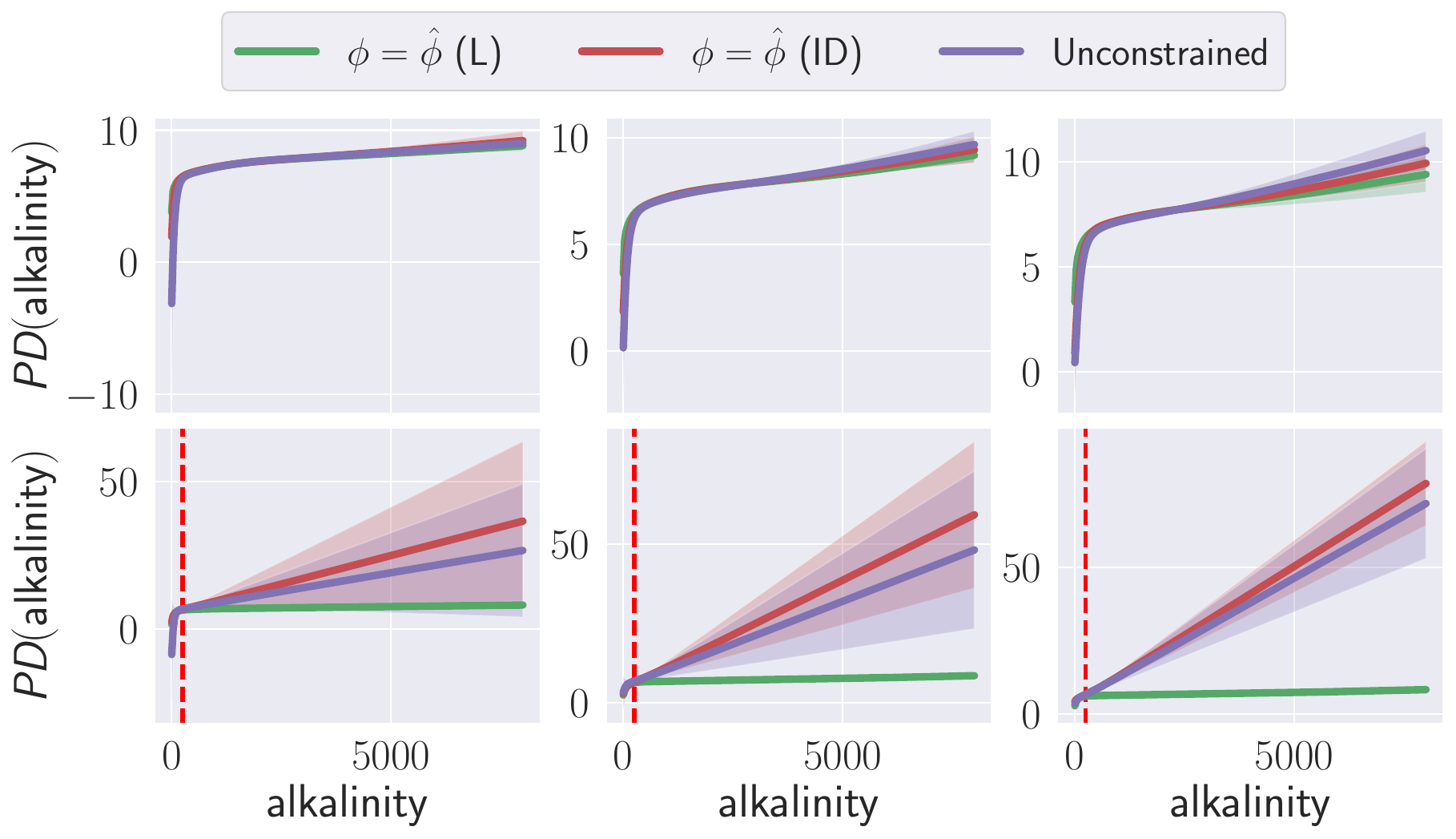}
    \caption{The figure displays the mean PD estimations for $\textit{PD}_N(h^{\theta}, \text{alkalinity})$ obtained with each method, on the PHALK dataset, in the interpolation (top) and extrapolation settings (bottom). All methods have been trained respectively on 100 samples (left), 50 samples (middle), 25 samples (right). Mean and standard deviations are computed over 20 different dataset splits. The dotted red lines represent the boundaries of the interpolation range.}
    \label{fig:phalk_all_sizes_pd_comparison}
\end{figure}

In Figure \ref{fig:phalk_all_sizes_pd_comparison}, we provide the mean PD estimations, both in the interpolation (top) and extrapolation (bottom) settings. In the interpolation setting, we observe that the PD estimations of all models follow the same logarithm-like trend, for all training sizes. 
In the extrapolation setting, the situation is quite different. As was the case for the previous experiment, both unconstrained models and in-distribution constraining do not manage to recover the logarithmic structure outside the training domain, which is verified with their respective performance values in Figure \ref{fig:phalk_mse_interpolation_extrapolation} (right). Linspace constraining, on the contrary, helps to better model the shape of the desired PD, thereby improving final performance.


\subsection{Comparison to physics-informed models}
\label{subsec:comparison_phy_informed}
We provide in Appendix \ref{app:sec_phy_informed_comparison} a comparison between the different constraining methods and two physics-informed training methods, namely APHYNITY \citep{yin2021augmenting} and partial-dependence-based hybrid modeling \citep{claes2025hybrid}. Although the goal of both approaches is radically different from ours, we deemed interesting to have a look at this specific type of methods in terms of prediction performance but mostly in terms of PD estimation performance, as they are straightforward to implement for the considered nature of knowledge. 

In short, we show that unless the considered problem falls into the assumptions made by such methods (e.g., additive decomposition), then constraining methods, and especially linspace constraining, generally yield better-performing models that are more robust to out-of-distribution samples and whose PD estimates are more faithful to the prior knowledge.

\subsection{Dynamical systems forecasting}
\label{subsec:dyn_systems}
Our algorithm can be extended to the problem of dynamical systems forecasting. In Appendix \ref{app:subsec_dyn_systems}, we derive the corresponding formalism and provide results for the damped pendulum system, for which we show that constrained models outperform their unconstrained counterparts, in particular when the number of trajectories becomes scarcer and for trajectories generated from out-of-distribution initial states.

\subsection{Constraining with mis-specified knowledge}
In Appendix \ref{app:misspecified_prior}, we study the performance of model constraining with mis-specified prior knowledge. The idea is to evaluate the impact of prior modeling errors in the expected PD on the performance of constrained models. In short, we show that our method can be robust to mis-specified knowledge provided that the latter still reasonably approximates the actual true prior knowledge. Nevertheless, it is striking that even a vague prior helps to drastically reduce the performance drop in extrapolation with linspace-constrained models, compared to unconstrained models.

\section{Conclusion}
We introduce a new approach in EGL, based on PD functions, to constrain the training of neural networks. We compare different constraining approaches, depending on the level of available knowledge. On several regression problems, including dynamical systems forecasting and real-world datasets, we empirically show that our constraining methods help to learn models that produce more reliable PD estimates, but also models with stronger predictive performance. In particular, linspace constraining approaches generally perform much better than both in-distribution training and unconstrained training. 
Finally, we show that linspace constraining yields robust neural networks when observing test-time domain shift, even with very scarce training data or with mis-specified prior knowledge. We also show that compared to physics-informed hybrid models, our constrained models generally perform better and provide more faithful PD estimates, especially in out-of-distribution settings, as they are more generic and do not explicitly encode additivity.

As shown on the product problem, with correlated input features, we think that our method could be a way to improve the faithfulness of the resulting PDP of the model, in particular when using linspace constraints, as this will force the model to use the features involved in the PDP $(\x_k)$ instead of their correlated counterparts.

Our constraining method suffers from larger computing times compared to unconstrained neural networks, due to the estimation of $\textit{PD}_N(h^\theta, \x_k)$ which in principle requires the evaluation of the model on all N possible pairs $(\x_k, \x_{-k}^{(i)})_{i=1}^N$. It is nevertheless straightforward to reduce the number of data points by randomly sampling $n < N$ points from the original dataset, which would potentially come at the cost of worse partial dependence estimations of $h^\theta$, but it still remains an unbiased estimator and would reduce training time. Moreover, given that we target mostly low-data regimes, computing times is not expected to be a serious limitation.

As future work, we are interested in testing other methods to learn the initial prior function when $\phi = \hat{\phi}$. Indeed, we have currently only implemented this learning process through a complete fit of $h^\phi_k$ on the raw output but it might not always be an optimal choice, and the approximation of the PD function is crucial as it would most likely degrade model performance if fitted poorly. Another interesting problem would be to assume no explicit expert knowledge about the shape of the PD function, but rather to rely on an approximation through some fitted model.

Furthermore, our method can in principle be adapted to any other functional interpretation method, e.g., individual conditional expectation (ICE) measures. Therefore, we would like to investigate the impact of constraining with such measures on both predictive performance and explanation quality, which could help in problems with heterogeneous effects that the PD would cancel. 

Finally, beyond neural networks, it would be also interesting to adapt the framework to other popular regression methods such as boosting algorithms.

\subsubsection*{Broader Impact Statement}
This paper presents work whose goal is to advance the field
of Machine Learning. There are many potential societal
consequences of our work, none of which we feel must be
specifically highlighted here.

\bibliography{main}
\bibliographystyle{tmlr}

\appendix
\section{Additional results}

\subsection{Friedman problem}
\label{app:friedman}
Let us recall the problem definition:
\begin{align*}
    y = \psi_{0} \sin(\psi_{1} x_0 x_1) + \psi_{2} (x_2 - \psi_{3})^2 + \psi_{4} x_3+ \psi_{5} x_4 + \sum_{j=5}^{9}0 x_j + \varepsilon,
\end{align*}
where $x_j \sim \mathcal{U}(0, 1), j=0, \dots 9$, $\psi = [10, \pi, 20, 0.5, 10, 5]$ and $\varepsilon \sim \mathcal{N}(0, 1)$ \citep{friedman1983multidimensional}. We now consider the following two PD functions:
\begin{align*}
    h_k^{\phi}(x_2) &= \; 
        \phi_0 [\phi_{2} (x_2 - \phi_{3})^2] + \phi_1,\\
    h_k^{\phi}(x_3, x_4) &= \; \phi_0 [\phi_{2} x_3 + \phi_{3} x_4] + \phi_1.
\end{align*}
Potential ground truth parameter vectors can be respectively $\phi^* = [1, C_2, \psi_2, \psi_3]$ and $\phi^* = [1, C_3, \psi_4, \psi_5]$, where:
\begin{align*}
    C_2 &= \mathbb{E}_{\x_{-k}}\left[\psi_{0}\sin(\psi_{1} x_0 x_1) + \psi_{4} x_3 + \psi_{5} x_4\right],\\
    C_3 &=  \mathbb{E}_{\x_{-k}}\left[\psi_{0}\sin(\psi_{1} x_0 x_1) + \psi_{2} (x_2 - \psi_{3})^2\right].
\end{align*}

\subsubsection{Constraining $\textit{PD}_N(h^{\theta}, x_2)$}
\label{app:friedman_second_problem}

\paragraph{Sample size effect}

\begin{figure}[ht]
    \centering
    \includegraphics[width=\linewidth]{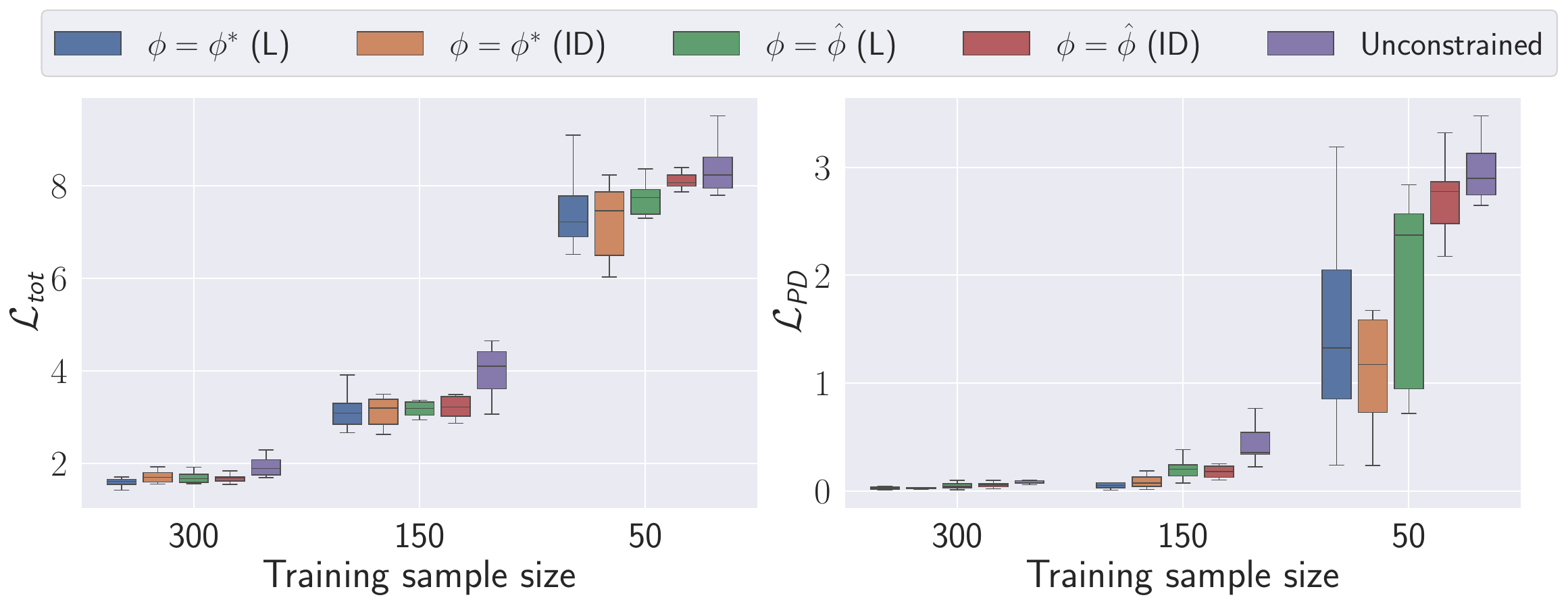}
    \caption{Evolution of $\mathcal{L}_{\textit{tot}}$ (left) and $\mathcal{L}_\textit{PD}$ (right) w.r.t. the training sample size, on the Friedman problem, constraining $\textit{PD}_N(h^{\theta}, x_2)$. Each boxplot summarizes the results over 10 different training initializations.}
    \label{fig:friedman_second_mse_pd_mse}
\end{figure}

We can observe in Figure \ref{fig:friedman_second_mse_pd_mse} that constraining $\textit{PD}_N(h^{\theta}, x_2)$ helps to improve both measures of performance, for each training size, and it now seems that linspace constraining and in-distribution constraining are mostly on par, which was not the case when constraining on $\textit{PD}_N(h^{\theta}, x_0, x_1)$. This may be explained by the fact that the PD function is now only one-dimensional and data points are uniformly sampled in $[0, 1]$, meaning that the input domain on $x_2$ is more densely covered than that of two-dimensional PD functions. Furthermore, we also hypothesize that both constraining methods (i.e., linspace and in-distribution) perform similarly because the PD shape is symmetrical and less intricate to approximate.

\begin{figure}[ht]
    \centering
    \includegraphics[width=\linewidth]{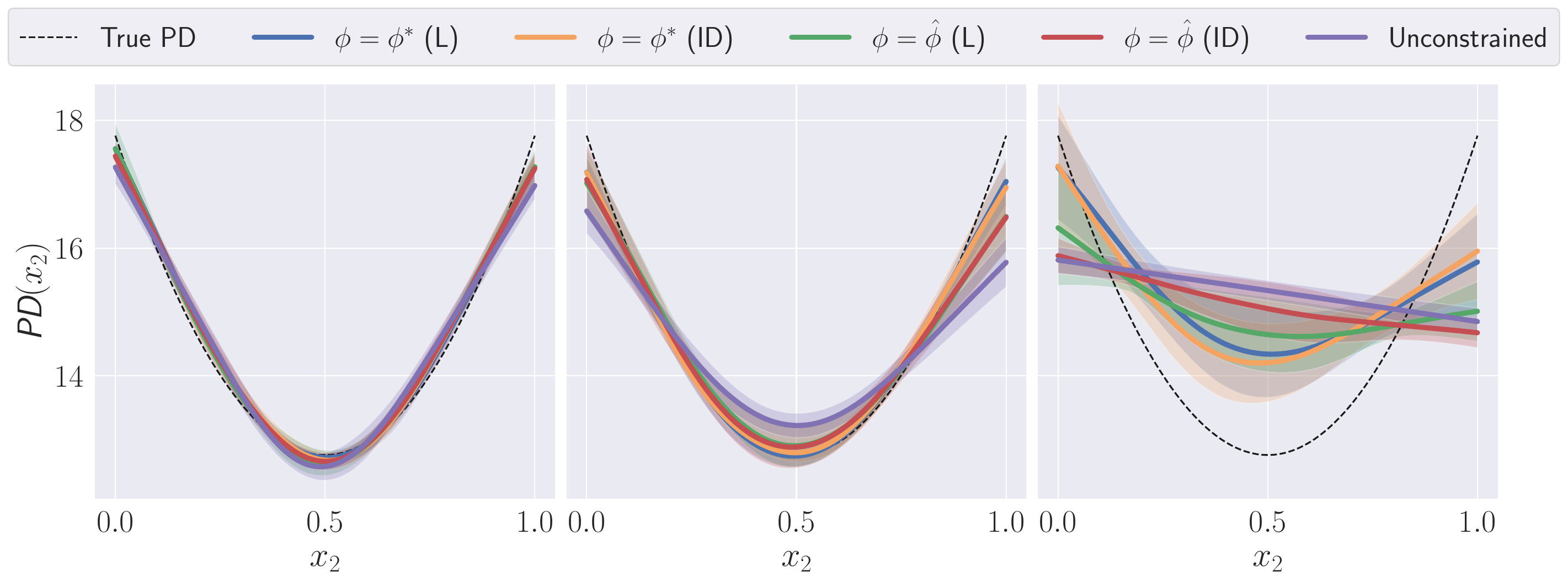}
    \caption{The figure displays the mean PD estimations for $\textit{PD}_N(h^{\theta}, x_2)$ obtained with each method, for the Friedman problem, in the interpolation setting. All methods have been trained respectively on 300 samples (left), 150 samples (middle), 50 samples (right). Mean and standard deviations are computed over 10 different training initializations.}
    \label{fig:friedman_second_pd_comparison_interpolation}
\end{figure}


    

Figure \ref{fig:friedman_second_pd_comparison_interpolation} displays the average PD estimations obtained when constraining $\textit{PD}_N(h^{\theta}, x_2)$ for each training size, in interpolation. We can observe that with enough data points (300), all methods (including unconstrained training) are able to provide reliable estimates. With half as many training samples, unconstrained PD estimates are now less precise, whereas with 50 training samples, it produces a straight curve. We also note that for the latter size, all training methods seem to shift the PD estimates upwards, which we attribute to the initial objective of minimizing \eqref{eq:sample_mse_def} overtaking the minimization of \eqref{eq:sample_pd_mse}, as only the former objective is used as model selection criterion.


\paragraph{Test-time domain shift}
We now enforce test-time domain shift in $x_2$. As was the case for the previous Friedman experiment, samples for which $x_2 > 0.75$ are rejected from the training and validation sets.

\begin{figure}[ht]
    \centering
    \includegraphics[width=\linewidth]{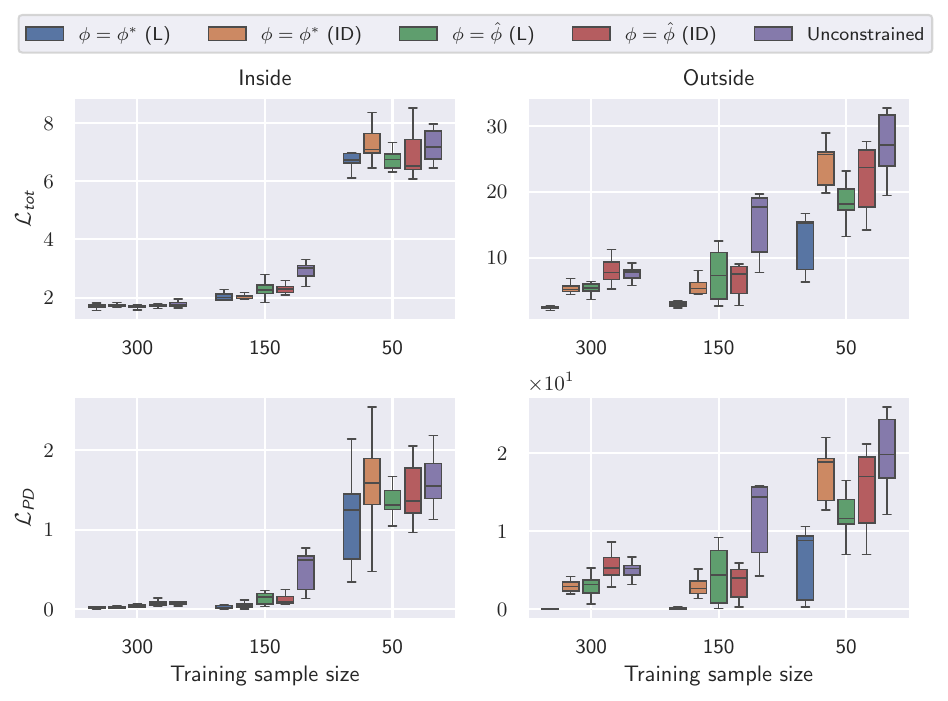}
    \caption{Evolution of $\mathcal{L}_{\textit{tot}}$ (top plots) and $\mathcal{L}_{\textit{PD}}$ (bottom plots) w.r.t. the training sample size, on the Friedman problem, constraining $\textit{PD}_N(h^{\theta}, x_2)$. \textit{Inside} metrics are computed on samples for which $x_2 \leq 0.75$ (left plots), while \textit{Outside} metrics are computed on samples where $x_2 > 0.75$ (right plots). Each boxplot summarizes the results on a test set, over 10 different training initializations.}
    \label{fig:friedman_second_inside_outside}
\end{figure}

\begin{figure}[ht]
    \centering
    \includegraphics[width=\linewidth]{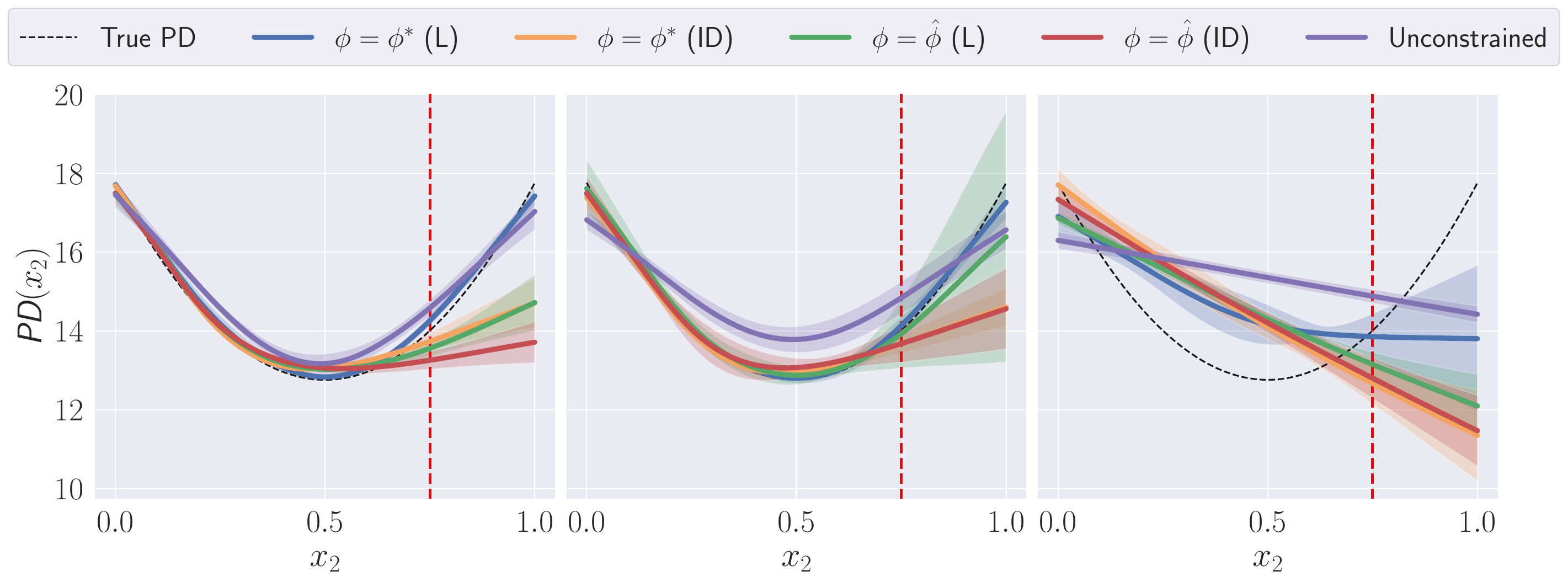}
    \caption{The figure displays the mean PD estimations for $\textit{PD}_N(h^{\theta}, x_2)$ obtained with each method, for the Friedman problem, in the extrapolation setting. All methods have been trained respectively on 300 samples (left), 150 samples (middle), 50 samples (right). Mean and standard deviations are computed over 10 different training initializations. The dotted red line represents the boundary of the interpolation range.}
    \label{fig:friedman_second_pd_comparison_extrapolation}
\end{figure}


    

In Figure \ref{fig:friedman_second_inside_outside}, we observe very similar results to those obtained in Figure \ref{fig:friedman_second_mse_pd_mse}, where constrained training improves both measures of performance. Nevertheless, we notice that after training on 50 samples, estimations of the PD function for both unconstrained training (purple boxes) and constrained training with an approximated prior (green and red boxes) seem better than previously, by simply looking at Figure \ref{fig:friedman_second_inside_outside} (left). This can however be explained by the fact that the target domain is now cut at $x_2 = 0.75$ for \textit{inside} estimates, therefore omitting an important part where larger error amplitudes are concentrated. This can be further verified by analyzing Figure \ref{fig:friedman_second_pd_comparison_extrapolation} (right). Indeed, we can see that linspace constraining on the ground truth PD function yields the least worse estimations, as it is the only method that has access to high-quality information about the PD outside the interpolation domain. Even though the optimal in-distribution constraining (orange curve) also has access to the same level of quality, it is nonetheless limited up to $x_2 = 0.75$. As far as constraining with approximated priors is concerned (green and red curves), neither method produces good PD estimates as they are limited by the quality of the prior, which is only fitted inside the interpolation range. Furthermore, due to the limited amount of training points, it seems that, even in the interpolation range, models put more weight to the minimization of the total mean-squared error rather than the minimization of the distance to the (true) target PD function.

For comparison, we display in Figure \ref{fig:friedman_second_pd_comparison_extrapolation} (left) the PD estimations obtained in extrapolation, after training with 300 samples. We can observe that now all methods provide quite reliable PD estimates in the interpolation domain, which was not the case with only 50 samples. Furthermore, linspace constraining on the true PD function (blue curve) vastly outperforms other baselines, which is expected. It is interesting to notice that, in this scenario, linspace constraining (green curve) produces PD estimations similar to in-distribution constraining on the true PD function (orange curve). This means that, even with imperfect knowledge about the latter, it still seems better to constrain on a wider region. Even more interestingly, while estimations after training on 150 samples show more variance, it seems that mean PD estimations of such linspace fitting are closer to the true PD function than mean estimations obtained on 300 samples. PD estimates obtained from unconstrained training clearly degrade as training size decreases.

\subsubsection{Constraining $\textit{PD}_N(h^{\theta}, x_3, x_4)$}
\label{app:friedman_third_problem}

\begin{figure}[ht]
    \centering
    \includegraphics[width=\linewidth]{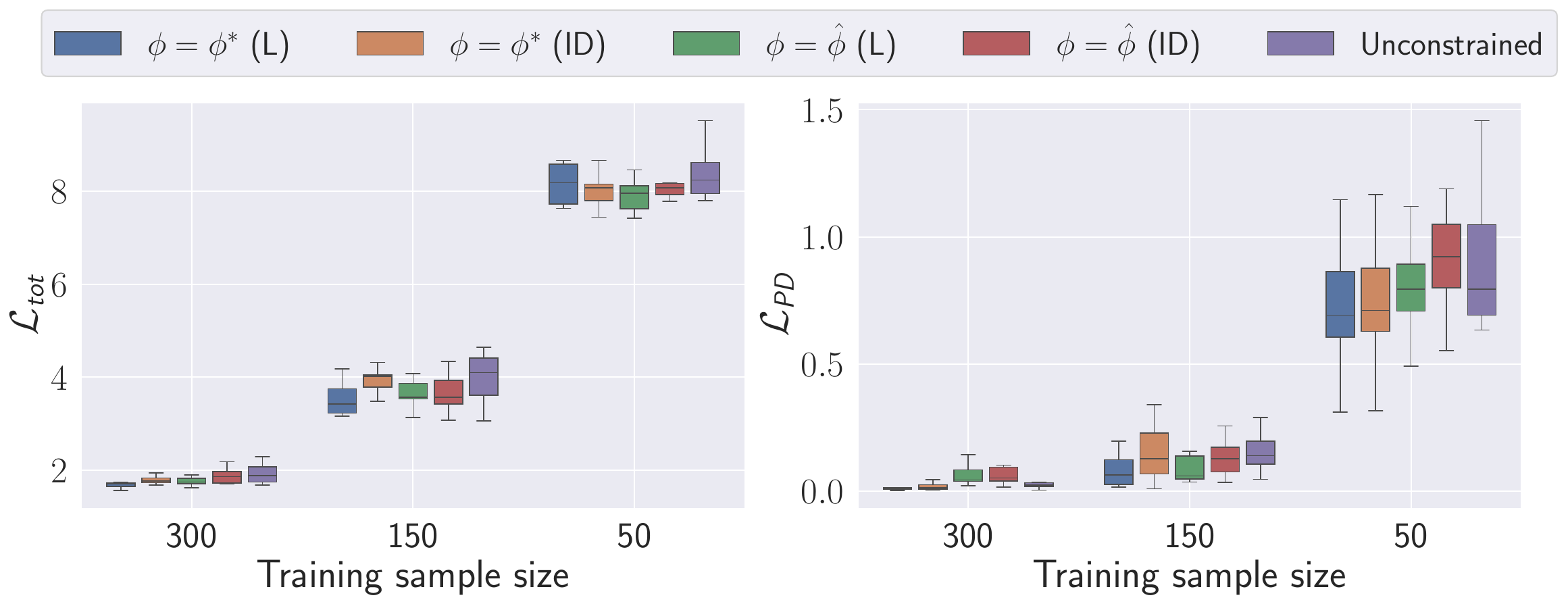}
    \caption{Evolution of $\mathcal{L}_{\textit{tot}}$ (left) and $\mathcal{L}_{\textit{PD}}$ (right) w.r.t. the training sample size, on the Friedman problem, constraining $\textit{PD}_N(h^{\theta}, x_3, x_4)$. Each boxplot summarizes the results over 10 different training initializations.}
    \label{fig:friedman_third_mse_pd_mse}
\end{figure}

\begin{figure}[ht]
    \centering
    \includegraphics[width=\linewidth]{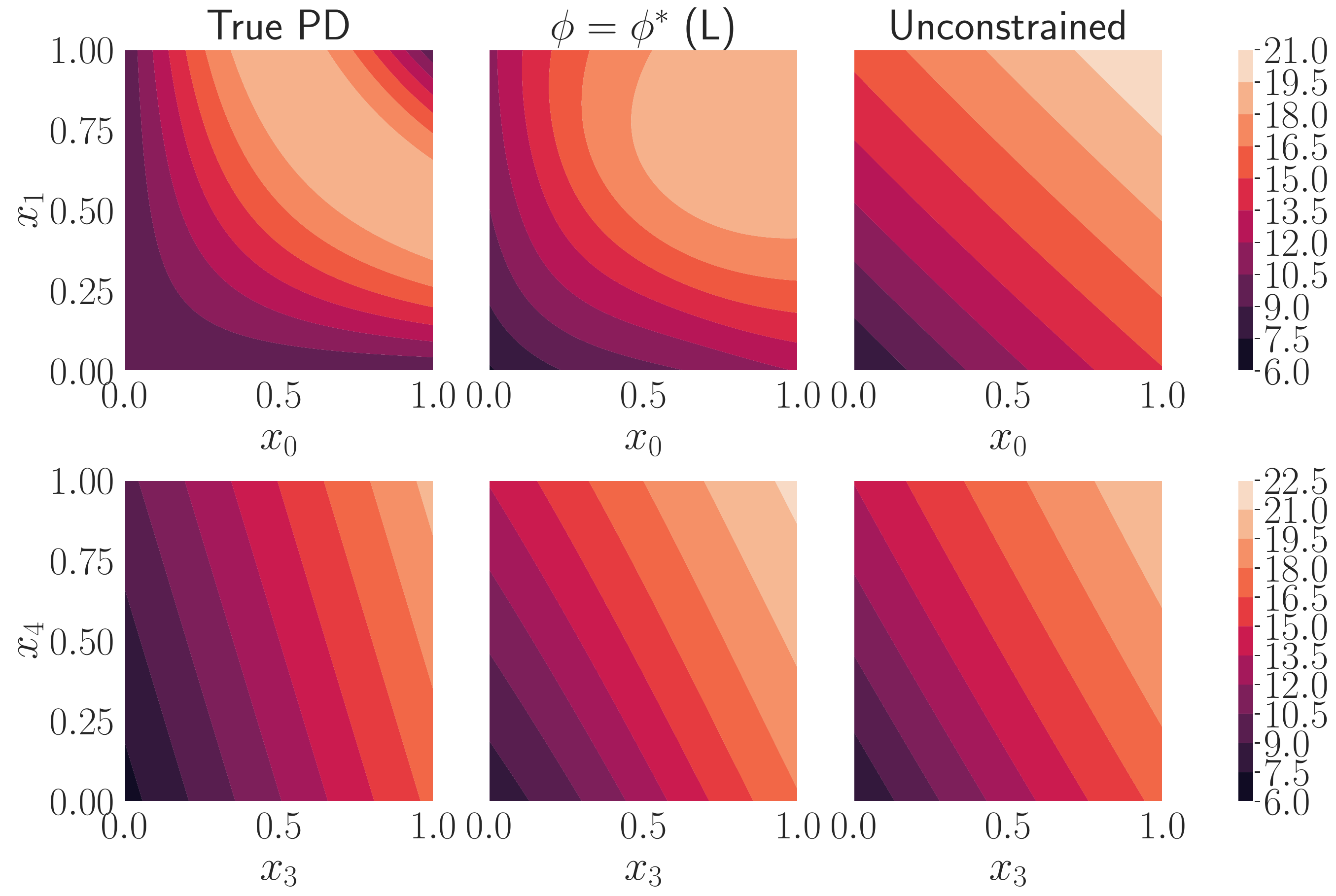}
    \caption{Example of PD estimation for $\textit{PD}_N(h^{\theta}, x_0, x_1)$ (top plots) and $\textit{PD}_N(h^{\theta}, x_3, x_4)$ (bottom plots). The figure displays the true PD (left), the PD after linspace constraining on the ground truth PD (middle), and the PD after unconstrained training (right). Both have been trained with 50 samples, in the interpolation setting.}
    \label{fig:pd_first_third_50_samples}
\end{figure}

\paragraph{Sample size effect}
While the impact of PD constraining was quite marked for $\textit{PD}_N(h^{\theta}, x_0, x_1)$ and $\textit{PD}_N(h^{\theta}, x_2)$, constraining on $\textit{PD}_N(h^{\theta}, x_3, x_4)$ seems to have very little impact on both the total mean-squared error and on the estimated PD, as can be seen in Figure \ref{fig:friedman_third_mse_pd_mse}. In fact, we notice that the scale of the PD error is much smaller than that of the two previous experiments, floating around $0.8$ for 50 samples, whereas it is twice to five times as large for $\textit{PD}_N(h^{\theta}, x_0, x_1)$ and $\textit{PD}_N(h^{\theta}, x_2)$. This null impact of PD constraining can be explained by the fact that $h_k^{\phi}(x_3, x_4) = \; \phi_0 [\phi_{2} x_3 + \phi_{3} x_4] + \phi_1$ is actually very easy to capture from the output signal. To illustrate this claim, we provide in Figure \ref{fig:pd_first_third_50_samples} samples of PD estimations for $\textit{PD}_N(h^{\theta}, x_0, x_1)$ (top) and $\textit{PD}_N(h^{\theta}, x_3, x_4)$ (bottom), after training with 50 samples. It is easy to notice, for the latter, that even unconstrained training yields a PD estimation that is close to the real one, which is not true for the former.

\paragraph{Test-time domain shift}
As for $\textit{PD}_N(h^{\theta}, x_0, x_1)$, samples for which $x_3 > 0.75$ are rejected from the training and validation sets.

\begin{figure}[ht]
    \centering
    \includegraphics[width=\linewidth]{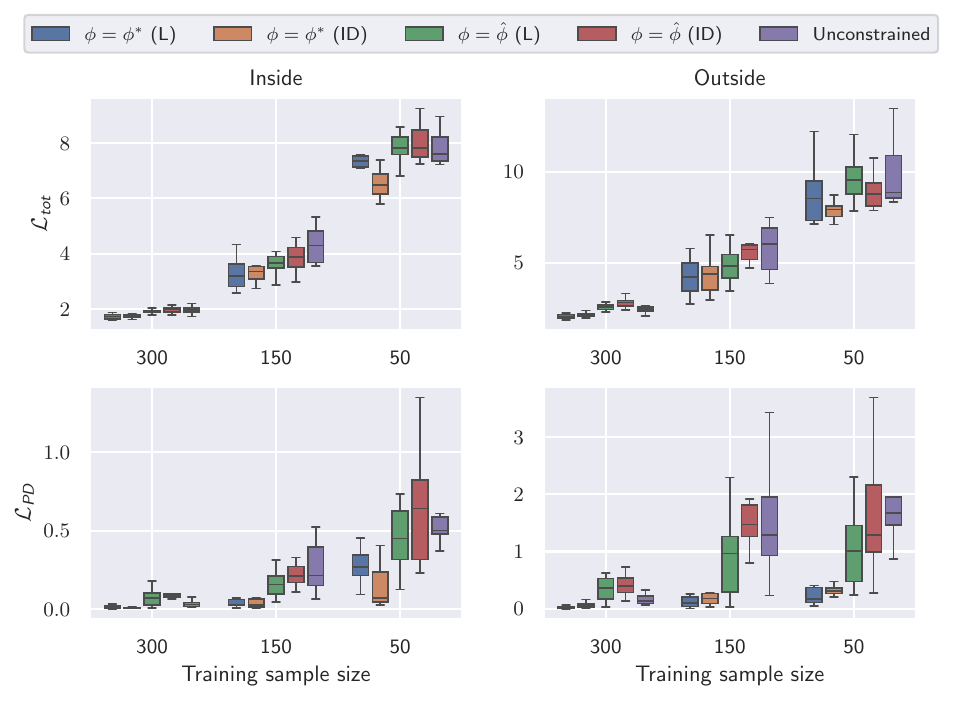}
    \caption{Evolution of $\mathcal{L}_{\textit{tot}}$ (top plots) and $\mathcal{L}_{\textit{PD}}$ (bottom plots) w.r.t. the training sample size, on the Friedman problem, constraining $\textit{PD}_N(h^{\theta}, x_3, x_4)$. \textit{Inside} metrics are computed on samples for which $x_3 \leq 0.75$ (left plots), while \textit{Outside} metrics are computed on samples where $x_3 > 0.75$ (right plots). Each boxplot summarizes the results on a test set, over 10 different training initializations.}
    \label{fig:friedman_third_inside_outside}
\end{figure}

\begin{figure}[ht]
    \centering
    \includegraphics[width=\linewidth]{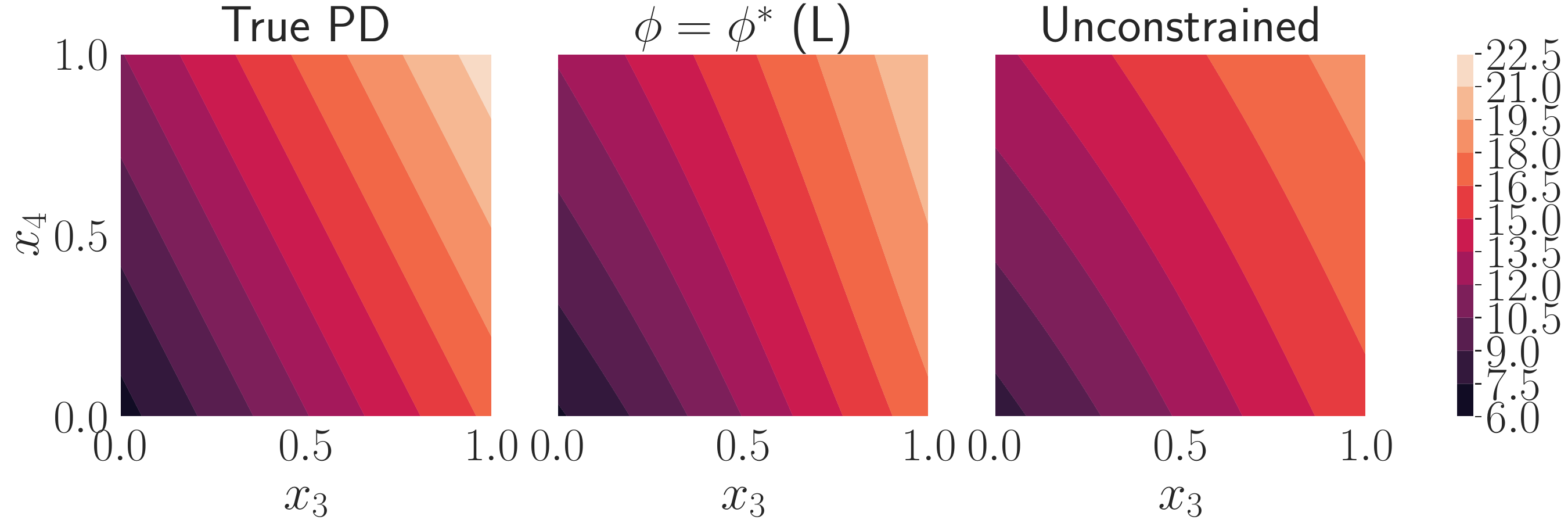}
    \caption{Example of PD estimation for $\textit{PD}_N(h^{\theta}, x_3, x_4)$. The figure displays the true PD (left), the PD after linspace constraining on the ground truth PD (middle), and the PD after unconstrained training (right). Both have been trained with 50 samples, in the extrapolation setting.}
    \label{fig:friedman_pd_third_50_samples_extrapolation}
\end{figure}

Similarly to the interpolation setting, we can observe in Figure \ref{fig:friedman_third_inside_outside} that constraining on $\textit{PD}_N(h^{\theta}, x_3, x_4)$ slightly helps in extrapolation, in terms of total mean-squared error. Nevertheless, the resulting PD estimations after constraining with the ground truth prior now seem to be consistently better than unconstrained training, which can be verified in Figure \ref{fig:friedman_pd_third_50_samples_extrapolation}. Even though both training methods captured the shape of the underlying PD function, unconstrained training fails to approximate accurately the output scale, especially outside the interpolation range, i.e. where $x_3 > 0.75$.

\subsection{Dynamical systems forecasting}
\label{app:subsec_dyn_systems}

Our algorithm can be straightforwardly extended to the problem of dynamical systems forecasting. In this problem, we are interested in predicting the evolution of a system driven by some dynamics $f$
\begin{equation}
\label{eq:dyn_system}
    \dot{\x}_t := \frac{d\x_t}{dt} = f(\x_t),
\end{equation}
where $\x_t \in \mathbb{R}^d$, and $f(\x_t)$ shares the same dimension as $\x_t$. Similarly, we assume knowledge about the dynamics function in the form of a PD function, and re-use previous model definitions of $h^{\theta}$ and $h_k^{\phi}$.

We assume access to a finite set $\textit{LS}$ of $N$ observed state trajectories over a given time interval $[0, T]$, where $T$ is the time horizon, i.e. $\textit{LS} = \{\x_{0}^{(i)}, \dots, \x_{T}^{(i)}\}_{i=1}^N$. We are still interested in solving \eqref{eq:final_objective}, i.e.
\begin{align*}
    \theta^*, \phi^* = \arg \min_{\theta, \phi} \mathcal{L}_{\lambda}(h^{\theta}, h_k^{\phi}, f; \textit{LS}),
\end{align*}
with
\begin{equation*}
    \mathcal{L}_{\lambda}(h^{\theta}, h_k^{\phi}, f; \textit{LS}) =  
         (1-\lambda) \mathcal{L}_{\textit{tot}}(h^{\theta}, f; \textit{LS})
    + \lambda \mathcal{L}_{\textit{PD}}(h^{\theta}, h_k^{\phi}; \textit{LS}),
\end{equation*}
for which
\begin{align}
    \label{eq:dyn_sample_pd}
    \textit{PD}_N(h^{\theta}, \x_{k}) &= \frac{1}{T}\frac{1}{N}\sum_{t=1}^{T}\sum_{i=1}^{N}h^{\theta}(\x_{k}, \x^{(i)}_{t,-k}),\\
    \label{eq:dyn_sample_global_mse}
    \mathcal{L}_{\textit{tot}}(h^{\theta}, f; \textit{LS}) &= \frac{1}{T}\frac{1}{N}\sum_{t=1}^{T}\sum_{i=1}^{N}\lVert\hat{\x}_t^{(i)} - \x_t^{(i)}\rVert^2,\\
    \label{eq:dyn_sample_pd_mse}
    \mathcal{L}_{\textit{PD}}(h^{\theta}, h_k^{\phi}; \textit{LS}, \textit{LS}_{\textit{PD}})&=
    \frac{1}{N_{\textit{PD}}}\sum_{j=1}^{N_{\textit{PD}}}(\textit{PD}_N(h^{\theta}, \x_{k}^{(j)}) - h_k^{\phi}(\x_{k}^{(j)}))^2,
\end{align}
where $\x_k$ is any value of the features of interest, $\hat{\x}_t^{(i)}$ is the solution of integrating $h^{\theta}$ with initial condition $\x_0^{(i)}$ up to time $t$, and $\textit{LS}_{\textit{PD}} = \{\x_k^{(j)}\}_{j=1}^{N_{\textit{PD}}}$ is a set of arbitrary values for the subset of features of interest.

As was the case before, we need to initialize the process, hence we arbitrarily fit $h_k^{\phi}$ for $E$ epochs on the raw trajectories, yielding
\begin{align*}
    \phi^{(0)} &= \arg\min_{\phi} \mathcal{L}_{\textit{prior}}(h_k^{\phi}, f; \textit{LS}),
\end{align*}
with
\begin{equation}
    \label{eq:dyn_sample_prior_mse}
    \mathcal{L}_{\textit{prior}}(h_k^{\phi}, f; \textit{LS}) = \frac{1}{T}\frac{1}{N}\sum_{t=1}^{T}\sum_{i=1}^{N}\lVert\hat{\x}_{t,k}^{(i)} - \x_{t,k}^{(i)}\rVert^2,
\end{equation}
where $\hat{\x}_{t,k}^{(i)}$ is the solution of integrating $h_k^{\phi}$ with initial condition $\x_{0,k}^{(i)}$ up to time $t$. 

\subsubsection{Damped pendulum system}
Let us consider the following dynamical system:
\begin{align*}
    \quad &\frac{d^2x_0}{dt^2} + \psi_0 \sin x_0 + \psi_1 \frac{dx_0}{dt} = 0, \\ \iff &\begin{cases}
        \dot{x}_0 = \frac{dx_0}{dt}\\
        \ddot{x}_0 = - \psi_0 \sin x_0 - \psi_1 \frac{dx_0}{dt}
    \end{cases}
\end{align*}
with $\x = (x_0, \frac{dx_0}{dt})$ the state, whose initial values are respectively sampled from $\mathcal{U}(-\pi, \pi)$ and $\mathcal{U}(0, \frac{1}{1000})$, and $\psi = [(\frac{\pi}{6})^2, 0.2]$, with the following PD function:
\begin{align*}
    h_k^{\phi}(x_0) = \phi_0[- \phi_2 \sin x_0] + \phi_1.
\end{align*}
For this problem, even though prior knowledge is only defined for $\ddot{x}_0$ through the PD, we still need a dynamical model for $x_0$ to track its evolution over time as it serves as the argument for the PD. Given that $x_1 = \frac{dx_0}{dt}$, we therefore redefine $h_k^{\phi} : (x_0, x_1) \rightarrow (x_1, \phi_0[- \phi_2 \sin x_0] + \phi_1)$.

A potential ground truth parameter vector is $\phi^* = [1, C, \psi_0]$, where:
\begin{equation*}
    C = \mathbb{E}_{\frac{dx_0}{dt}}\left[- \psi_1 \frac{dx_0}{dt}\right].
\end{equation*}

We generate training and validation sets of successively 50, 25 and 10 trajectories each. Each trajectory spans 81 evenly spaced timesteps (including the initial state $\x_0$), over a time interval of $[0, 20]$s, integrated through the Neural ODE approach \citep{chen2018neuralode}, using \texttt{dopri5} as solver. All methods are then evaluated on a fixed test set of 200 trajectories, and we set the batch size to 4 trajectories.

\begin{figure}[ht]
    \centering
    \includegraphics[width=\linewidth]{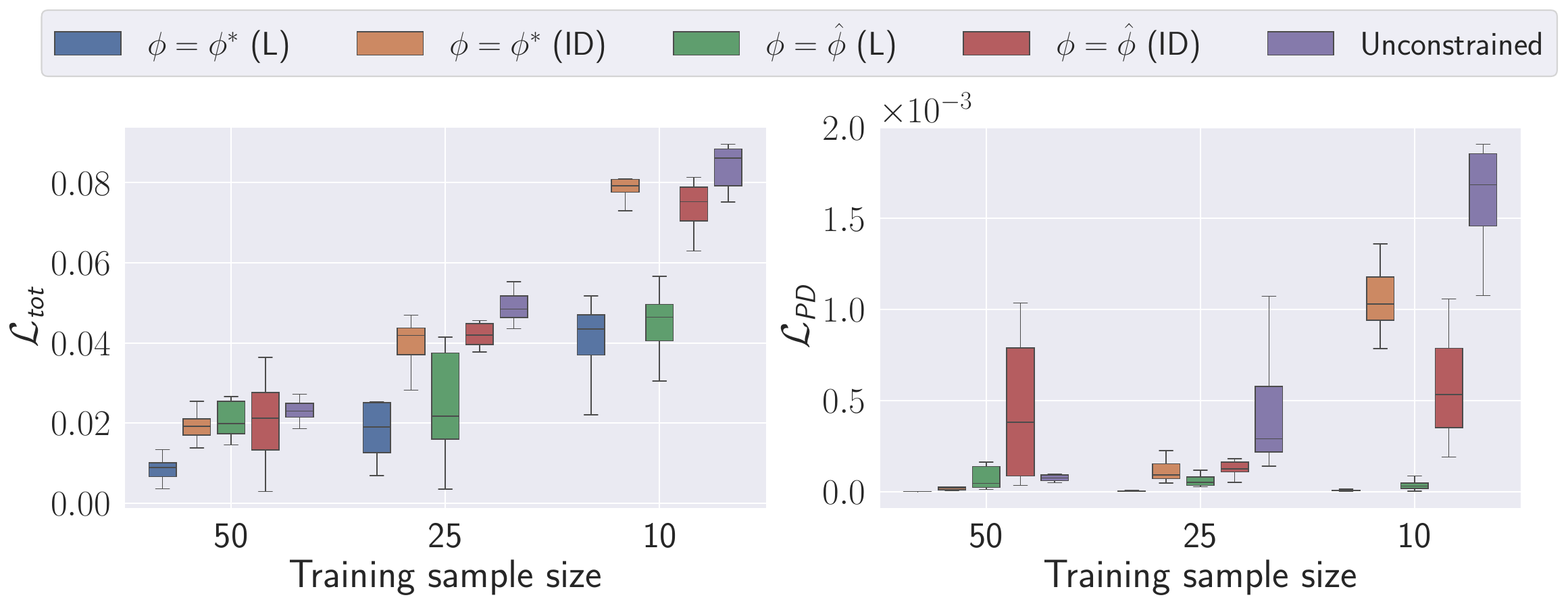}
    \caption{Evolution of $\mathcal{L}_{\textit{tot}}$ (left) and $\mathcal{L}_{\textit{PD}}$ (right) w.r.t. the training sample size, on the damped pendulum problem, constraining $\textit{PD}_N(h^{\theta}, x_0)$. Each boxplot summarizes the results on the test set, over 10 different training initializations.}
    \label{fig:damped_pendulum__mse_pd_mse}
\end{figure}

\paragraph{Sample size effect} As can be seen in Figure \ref{fig:damped_pendulum__mse_pd_mse},
while constraining seems to have moderate impact on the predictive performance for a learning sample of 50 trajectories, it nevertheless helps to predict more accurate trajectories for smaller training sample sizes, especially for linspace constraining methods (blue and green boxes). This positive impact is reflected on the quality of the learned PD function, for all constraining approaches, although more deeply for linspace constraining. Furthermore, linspace constraining on $h_k^{\phi^*}$ reaches similar predictive performance compared to unconstrained training but with only 25 training trajectories, compared to 50 for the latter. The same tendency can be observed for linspace constraining on an approximated prior, though it slightly suffers from the variance of the learned prior function. Finally, in-distribution constraining does not really improve predictive performance, but helps to learn a better PD function for smaller training sizes.

\paragraph{Test-time domain shift}
To study test-time shift, initial states whose angle $x_0 \notin [-\frac{3\pi}{4}, \frac{3\pi}{4}]$ are rejected from the training and validation sets.

\begin{figure}[ht]
    \centering
    \includegraphics[width=\linewidth]{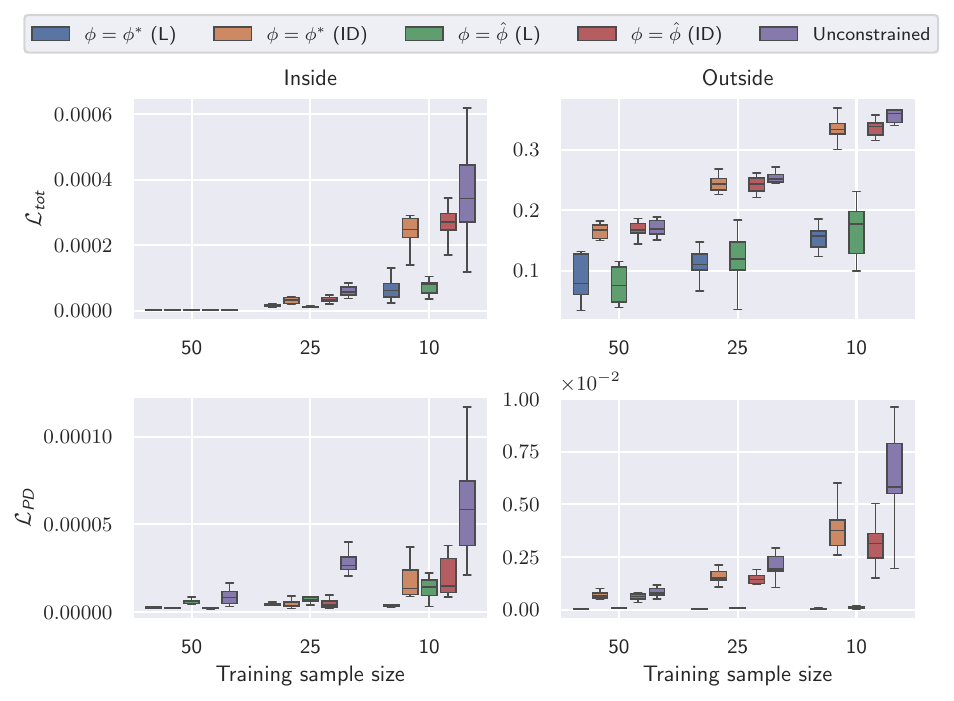}
    \caption{Evolution of $\mathcal{L}_{\textit{tot}}$ (top plots) and $\mathcal{L}_{\textit{PD}}$ (bottom plots) w.r.t. the training sample size, on the damped pendulum problem, constraining $\textit{PD}_N(h^{\theta}, x_0)$. \textit{Inside} metrics are computed on samples for which the initial $x_0 \in [-\frac{3\pi}{4}, \frac{3\pi}{4}]$ (left plots), while \textit{Outside} metrics are computed on samples outside that range (right plots). Each boxplot summarizes the results on the test set, over 10 different training initializations.}
    \label{fig:damped_pendulum_None_inside_outside_mse_and_pd_mse}
\end{figure}

We observe in Figure \ref{fig:damped_pendulum_None_inside_outside_mse_and_pd_mse} that predictive performance inside the interpolation range is now much lower for all training sizes and all methods, as a result of reducing the size of the input domain for the initial condition. As was the case previously, 50 training trajectories seem to be enough for unconstrained training to be as accurate \emph{inside} as constraining approaches, although producing slightly worse PD estimations. As the number of trajectories decreases, the gap between linspace approaches and unconstrained training becomes larger, and the former even perform on par with the latter with only 10 trajectories, compared to 25. Similarly to the previous experiment, in-distribution constraining does not really improve predictive performance (neither \emph{inside} nor \emph{outside} the interpolation range), but helps to learn better \emph{inside} PD estimates. However, it is striking that linspace constraining outperforms unconstrained training when starting from out-of-distribution initial states, both in terms of predictive performance and PD estimates.

\clearpage

\section{Comparison to physics-informed models}
\label{app:sec_phy_informed_comparison}

In this section, we compare the performance of our constraining methods against two physics-informed machine learning baselines, namely APHYNITY \citep{yin2021augmenting} and partial-dependence-based hybrid modeling \citep{claes2025hybrid}. Although radically different in terms of methodology, we deemed interesting to have a look at the performance that such methods could obtain for both measures of performance, i.e. $\mathcal{L}_{tot}$ and $\mathcal{L}_{\textit{PD}}$, on a series of problems.

\paragraph{APHYNITY}
Formally, APHYNITY combines a first-principles model, i.e. the prior knowledge $h_k(\x_k; \phi)$, with a ML model denoted $h_a(\x; \theta)$, in an additive manner, with learnable $\phi$ and $\theta$. The balance between $h_k$ and $h_a$ is controlled by minimizing the contribution of $h_a$ while trying to minimize the prediction errors of the entire model $h = h_k + h_a$, through the optimization of some regularization coefficient $\lambda \in \R$.

\paragraph{Partial-dependence-based hybrid modeling}
Similarly to APHYNITY, this hybrid method also additively combines the first-principles model $h_k$ with the ML model $h_a$. However, the optimization of both models is no longer performed jointly. Indeed, $h_k(\x_k, \phi)$ is first optimized by iteratively fitting the latter on PD estimations of an ML model. Then, $h_a(\x; \theta)$ is trained once on the residual $y - h_k(\x_k; \phi)$.

For all experiments, we fix the architecture of the neural network to three layers of increasing hidden size (32, 64 and 128), with a fixed training batch size of 16 samples (resp. 4 trajectories for dynamical systems and 5 for the real-world problems). We optimize the joint learning rate in $\{\num{5e-2}, \num{5e-3}, \num{5e-4}\}$ and the coefficient $\lambda$ in $\{1, 10, 20\}$, selecting the values that minimize $\mathcal{L}_{\textit{tot}}$ on the validation set. We set the number of epochs $E = 1000$. We refer to partial-dependence-based hybrid modeling as ``PD hybrid'' in the following results.

\subsection{Friedman problem}
Let us recall the problem definition:
\begin{align*}
    y = \psi_{0} \sin(&\psi_{1} x_0 x_1) + \psi_{2} (x_2 - \psi_{3})^2 + \psi_{4} x_3 + \psi_{5} x_4 + \sum_{j=5}^{9}0 x_j + \varepsilon,
\end{align*}
where $x_j \sim \mathcal{U}(0, 1), j=0, \dots 9$, $\psi = [10, \pi, 20, 0.5, 10, 5]$ and $\varepsilon \sim \mathcal{N}(0, 1)$ \citep{friedman1983multidimensional}.

We assume prior knowledge similar to our method, i.e.
\begin{align*}
    h_k(x_0, x_1; \phi) &= \; 
        \phi_{0}[\phi_2 \sin(\phi_{3} x_0 x_1)] + \phi_1,
\end{align*}
and the final model writes $h(\x) = h_k(x_0, x_1; \phi) + h_a(\x; \theta)$.

\begin{figure}[ht]
    \centering
    \includegraphics[width=\linewidth]{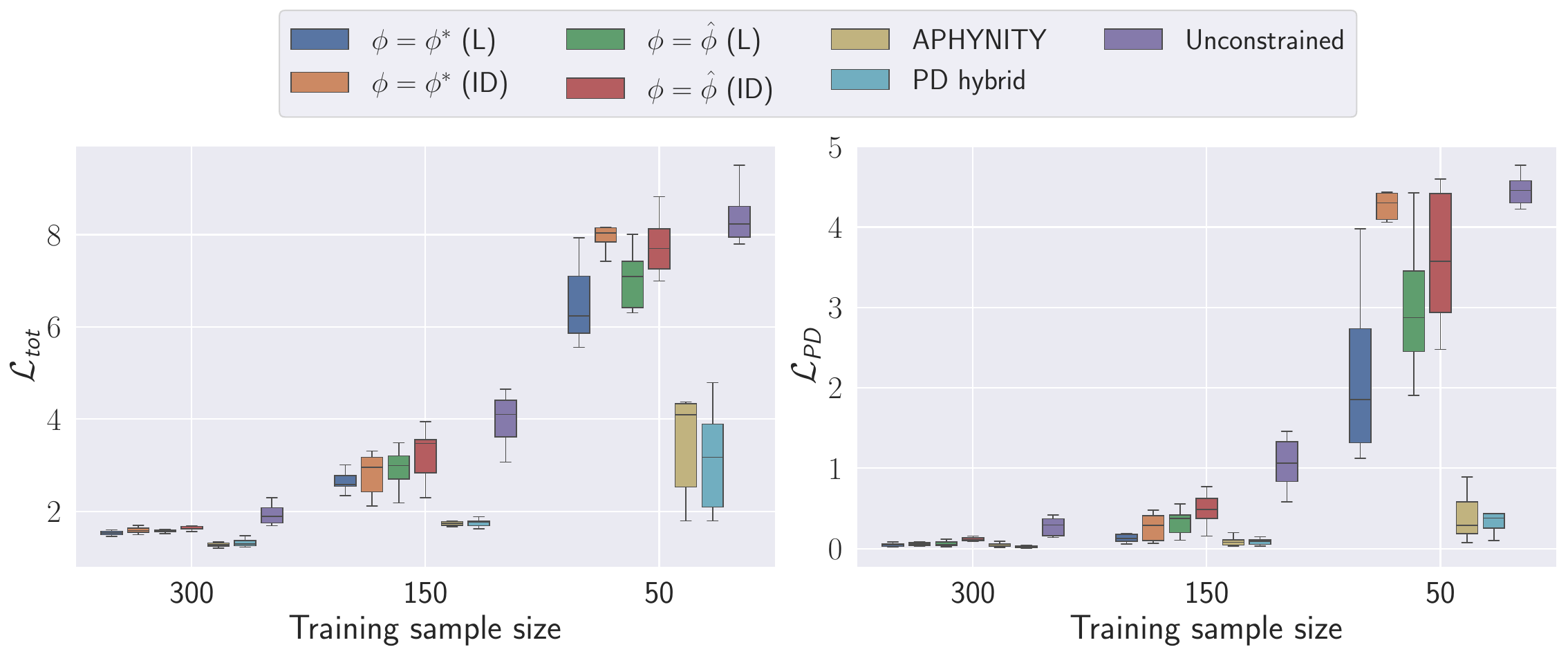}
    \caption{Evolution of $\mathcal{L}_{\textit{tot}}$ (left) and $\mathcal{L}_\textit{PD}$ (right) w.r.t. the training sample size, on the Friedman problem, constraining $\textit{PD}_N(h^{\theta}, x_0, x_1)$. Each boxplot summarizes the results over 10 different training initializations.}
    \label{fig:aphynity_friedman_first_mse}
\end{figure}

\paragraph{Sample size effect}
We observe in Figure \ref{fig:aphynity_friedman_first_mse} that hybrid methods perform the best for both measures of performance. This can actually be explained by the structure of this problem and by the inductive bias of hybrid additive methods. Indeed, given that this problem is purely additive with independent input features, it is therefore easy for such hybrid methods to converge towards faithful estimations of the prior knowledge. Note that this particular problem structure is also beneficial for constrained models with approximated priors, as the obtained priors are of better quality. Still, our method remains less restrictive as it does not assume model additivity but rather relies on a single model.

\begin{figure}[ht]
    \centering
    \includegraphics[width=\linewidth]{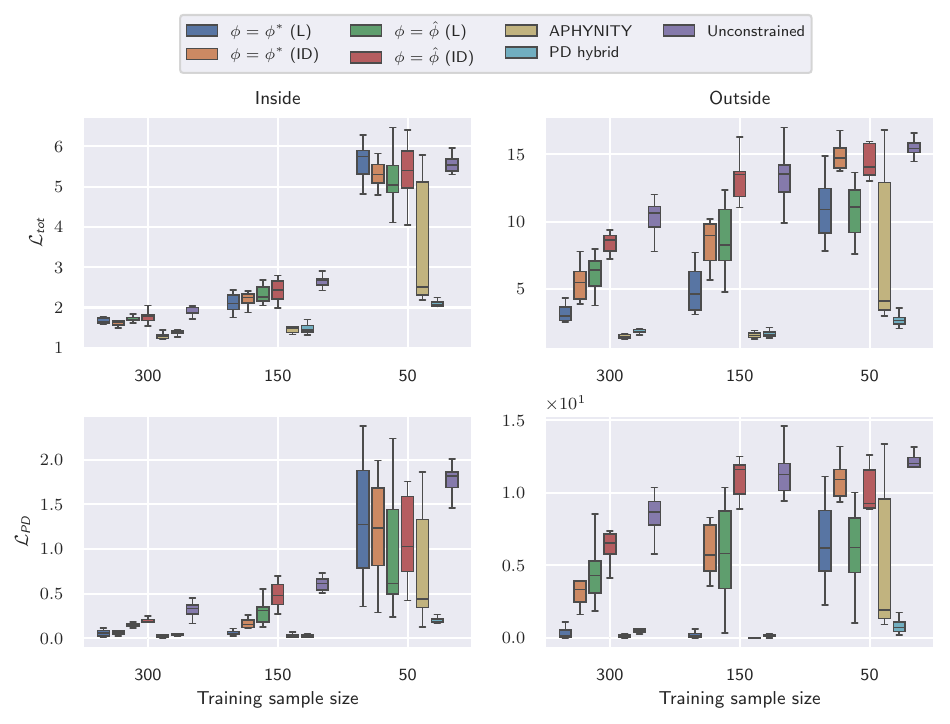}
    \caption{Evolution of $\mathcal{L}_{\textit{tot}}$ (top plots) and $\mathcal{L}_{\textit{PD}}$ (bottom plots) w.r.t. the training sample size, on the Friedman problem, constraining $\textit{PD}_N(h^{\theta}, x_0, x_1)$. \textit{Inside} metrics are computed on samples for which $x_0 \leq 0.75$ (left plots), while \textit{Outside} metrics are computed on samples where $x_0 > 0.75$ (right plots). Each boxplot summarizes the results on a test set, over 10 different training initializations.}
    \label{fig:aphynity_friedman_first_inside_outside}
\end{figure}

\paragraph{Test-time domain shift}
In Figure \ref{fig:aphynity_friedman_first_inside_outside}, we observe that hybrid methods are quite robust to test-time domain shift when training on 300 and 150 samples. We nevertheless notice that, when training on scarcer data (50 samples), the gap in performance on both measures is tightened between APHYNITY and linspace constraining, especially for $\mathcal{L}_{\textit{PD}}$. This could be explained by the fact that, due to the limited training domain, the quality of $h_k$ degrades and the statistical model $h_a$ then tries to compensate the misfit, which might lead to a worse PD. With linspace constraining, although the approximated prior also deteriorates, the constraint allows to regularize the shape of the PD outside the originally considered training domain. Still, we observe that PD-based hybrid modeling is not impaired by the extrapolation setting, which suggests that, for this problem, the iterative training procedure of $h_k$ on PD (residual) estimations of $h_a$ yields better physical parameters than those obtained with APHYNITY.

\subsection{Product problem}
Let us recall the problem definition:
\begin{align*}
    y = \psi_0 x_0^3 \cos(\psi_1 x_0) \frac{(x_1 - 1)^2}{2} + \sum_{j=2}^{9}0 x_j + \varepsilon,
\end{align*}
where features are correlated with zero mean, variances and covariances equal to $0.5$ and $0.25$, respectively, $\psi = [15, \pi]$, and $\varepsilon \sim \mathcal{N}(0, 1)$.

We assume prior knowledge through
\begin{equation*}
    h_k(x_0; \phi) = \phi_0 [\phi_2 x_0^3 \cos(\phi_3 x_0)] + \phi_1,
\end{equation*}
and the final model writes $h(\x) = h_k(x_0; \phi) + h_a(\x;, \theta)$.

\begin{figure}[ht]
    \centering
    \includegraphics[width=\linewidth]{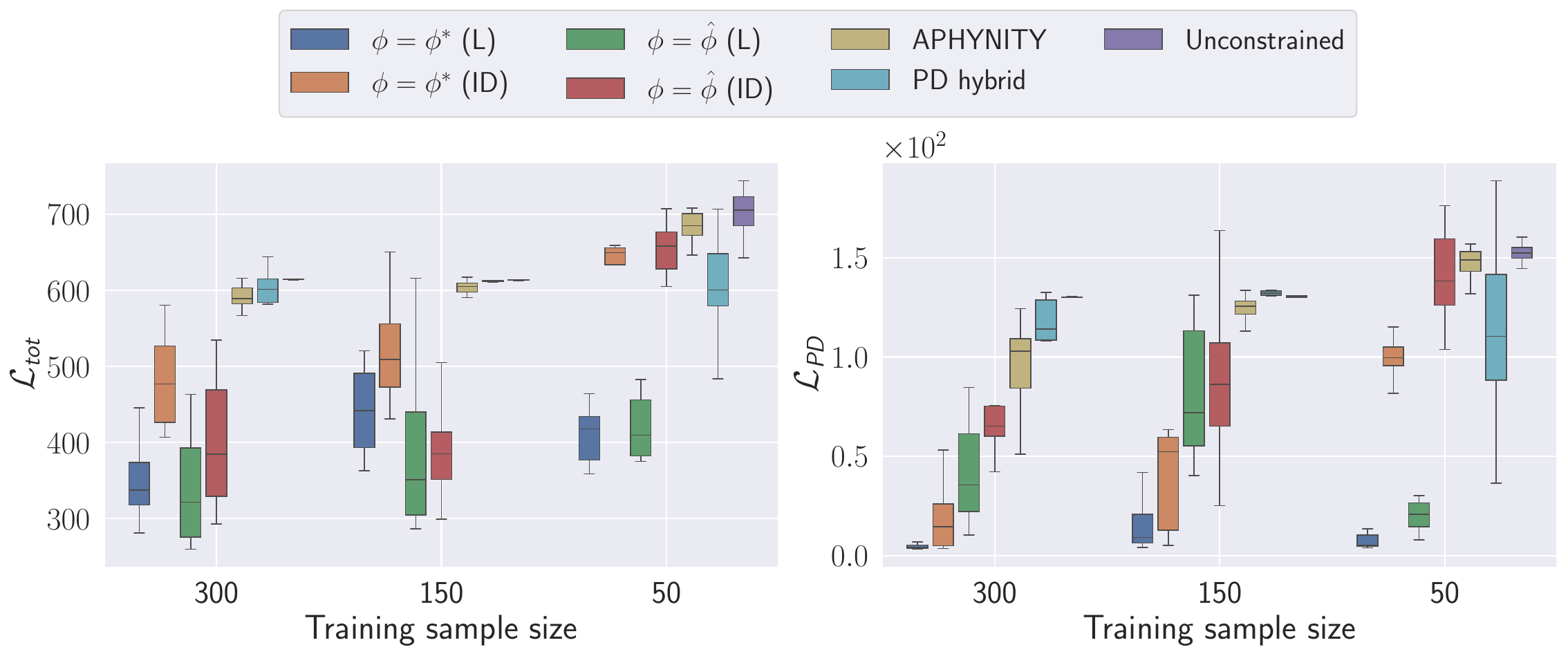}
    \caption{Evolution of $\mathcal{L}_{\textit{tot}}$ (left) and $\mathcal{L}_\textit{PD}$ (right) w.r.t. the training sample size, on the product problem, constraining $\textit{PD}_N(h^{\theta}, x_0)$. Each boxplot summarizes the results over 10 different training initializations.}
    \label{fig:aphynity_product_mse}
\end{figure}

\paragraph{Sample size effect}
We observe in Figure \ref{fig:aphynity_product_mse} that hybrid methods perform only slightly better than unconstrained models, for all training sizes.

This can be explained by the fact that such methods explicitly model additivity, which is a wrong assumption for this problem. We can notice that with sufficient data (300 samples), their PD estimation becomes slightly better than what would be obtained with an unconstrained neural network. On the contrary, our constraining method does not suffer from this particular problem structure, given that it does not encode additivity but simply relies on a single ML model. Although in-distribution constraining degrades with scarce data (50 samples) and performs roughly on par with hybrid models, linspace constraining vastly outperforms the other methods, for both measures.

\begin{figure}[ht]
    \centering
    \includegraphics[width=\linewidth]{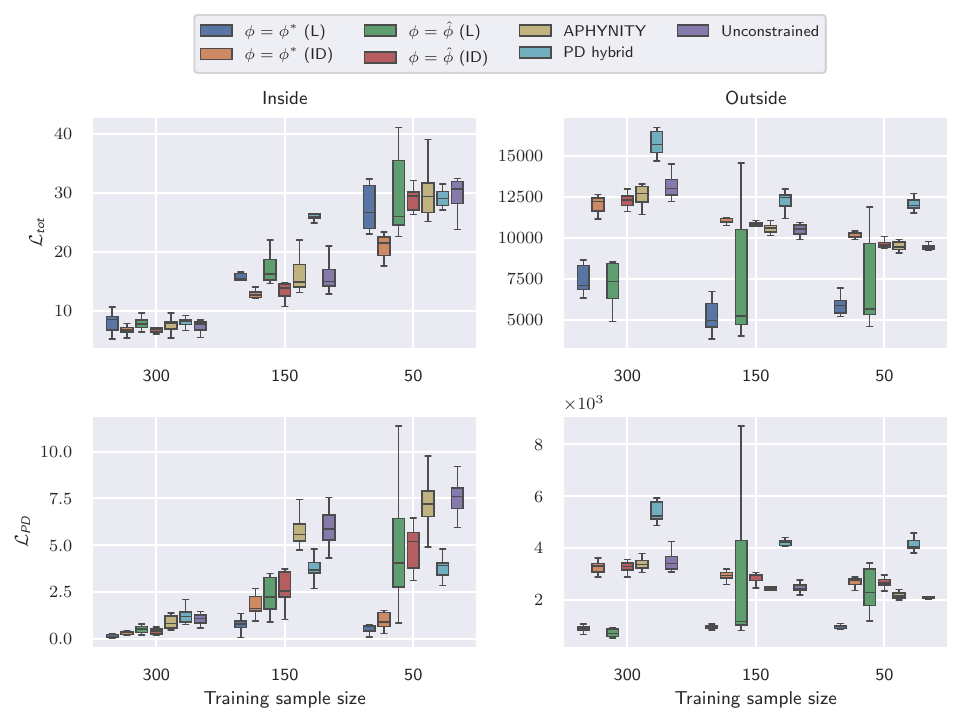}
    \caption{Evolution of $\mathcal{L}_{\textit{tot}}$ (top plots) and $\mathcal{L}_{\textit{PD}}$ (bottom plots) w.r.t. the training sample size, on the product problem, constraining $\textit{PD}_N(h^{\theta}, x_0)$. \textit{Inside} metrics are computed on samples for which $x_0 \in [-1.25, 1.25]$ (left plots), while \textit{Outside} metrics are computed on samples outside that range (right plots). Each boxplot summarizes the results on the test set, over 10 different training initializations.}
    \label{fig:aphynity_product_inside_outside}
\end{figure}

\paragraph{Test-time domain shift}
In Figure \ref{fig:aphynity_product_inside_outside}, we notice that even though the gap is closed inside the interpolation domain, their PD estimation still remains worse than that of constrained models, and equivalent to that of unconstrained networks.

For samples outside the interpolation range, we can observe that linspace constraining also outperforms hybrid models.

\subsection{Concrete compressive strength prediction}
As a reminder, the dataset is composed of samples relating characteristics of concrete components with the associated compressive strength. We assume similiar logarithmic knowledge, i.e.
\begin{align*}
    h_k(\text{age}; \phi) = \phi_0\ln(\text{age}) + \phi_1,
\end{align*}
and the final model writes $h(\x) = h_k(\text{age}; \phi) + h_a(\x)$.

\begin{figure}[ht]
    \centering
    \includegraphics[width=\linewidth]{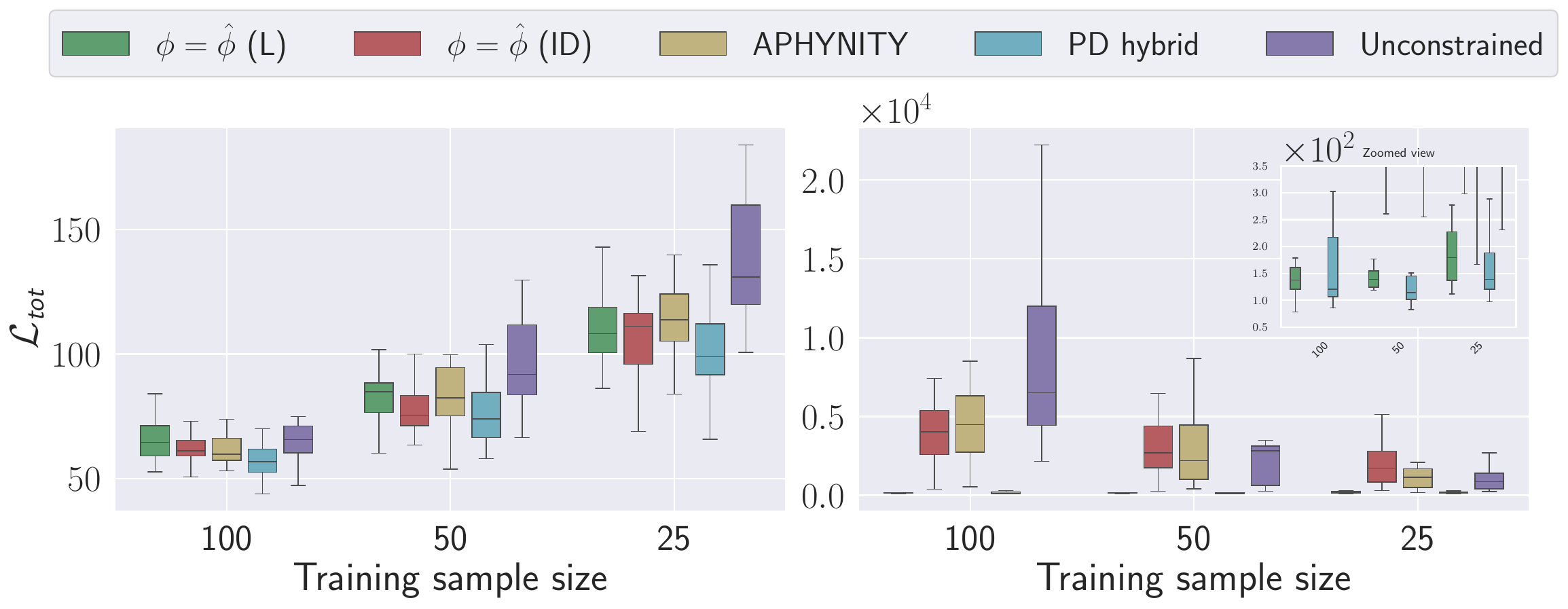}
    \caption{Evolution of $\mathcal{L}_{\textit{tot}}$ w.r.t. the training sample size, on the concrete problem, constraining $\textit{PD}_N(h^{\theta}, \text{age})$, in the interpolation setting (left) and extrapolation setting (right). Each boxplot summarizes the results on a test set, over 20 different dataset splits.}
    \label{fig:aphynity_concrete_mse_interpolation_extrapolation}
\end{figure}

\paragraph{Sample size effect}
We observe in Figure \ref{fig:aphynity_concrete_mse_interpolation_extrapolation} (left) that, as constrained models, hybrid methods reach better performance than unconstrained networks, for all training sizes. The gap with constrained methods is nonexistent, and we note that the performance of APHYNITY and linspace constraining are closely aligned. This can be explained by the fact that, for APHYNITY, the logarithmic knowledge holds for every value of age, by construction of the model. For linspace constraining, we enforce the PD to be logarithmic throughout the domain of age.

\paragraph{Test-time domain shift}
We observe in Figure \ref{fig:aphynity_concrete_mse_interpolation_extrapolation} (right) that APHYNITY falls short when facing out-of-distribution samples, performing roughly identically to in-distribution constraining and unconstrained models, meaning that, even though it is encoded in the model itself, it does not seem to be able to capture the logarithmic signal in the resulting PD, compared to linspace constraining. PD hybrid modeling performs much better than APHYNITY, but still slightly worse than linspace constraining with sufficient training samples, which suggests that the logarithmic effect of age could not be entirely additive, otherwise the gap would be tighter.

\begin{figure}[ht]
    \centering
    \includegraphics[width=\linewidth]{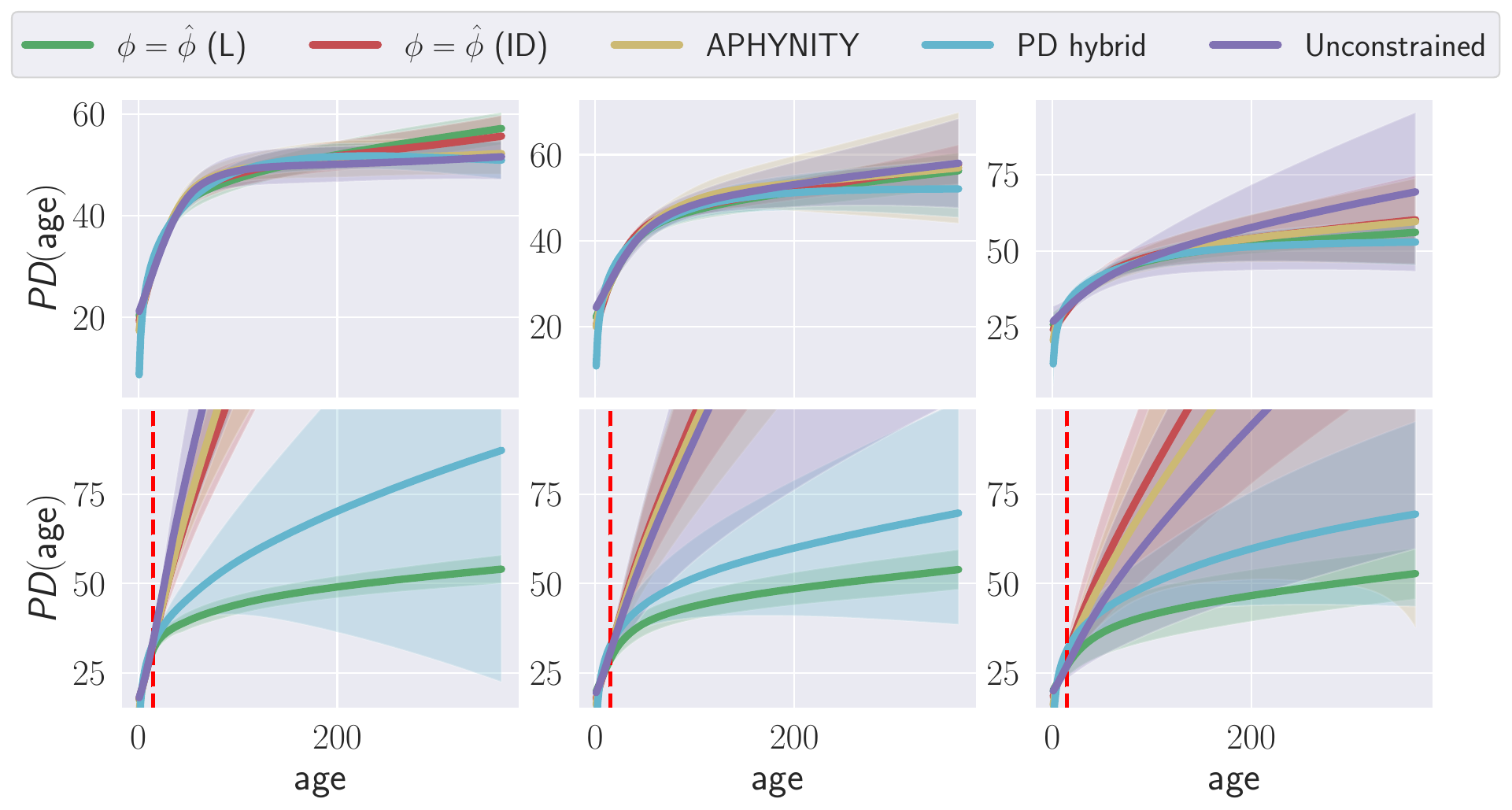}
    \caption{The figure displays the mean PD estimations for $\textit{PD}_N(h^{\theta}, \text{age})$ obtained with each method, for the compressive strength problem, in the interpolation (top) and extrapolation settings (bottom). All methods have been trained respectively on 100 samples (left), 50 samples (middle), 25 samples (right). Mean and standard deviations are computed over 20 different dataset splits. The dotted red lines represent the boundaries of the interpolation range.}
    \label{fig:aphynity_concrete_all_sizes_pd_comparison}
\end{figure}

We represent in Figure \ref{fig:aphynity_concrete_all_sizes_pd_comparison} the mean PD estimations, for all methods and training sizes. In particular, we can notice that in the interpolation setting (top), the PD estimations all follow a logarithmic shape. However, in the extrapolation setting (bottom), the PD estimations of APHYNITY completely diverge and follow the same trend as those of unconstrained models, showing that even though APHYNITY encodes the logarithmic relationship by construction, it is insufficient to result in a reliable PD. On the contrary, PD hybrid modeling suffers less than APHYNITY, as its PD estimation looks close to that of linspace constraining.

\subsection{PHALK dataset}
As a reminder, the dataset contains samples of basin characteristics, pH and alkalinity values, and we aim to predict pH values. We assume logarithmic knowledge, i.e.,
\begin{equation*}
    h_k(\text{alkalinity}; \phi) = \phi_0 \ln(\text{alkalinity}) + \phi_1,
\end{equation*}
and the final model writes $h(\x) = h_k(\text{alkalinity}; \phi) + h_a(\x)$.

\begin{figure}[ht]
    \centering
    \includegraphics[width=\linewidth]{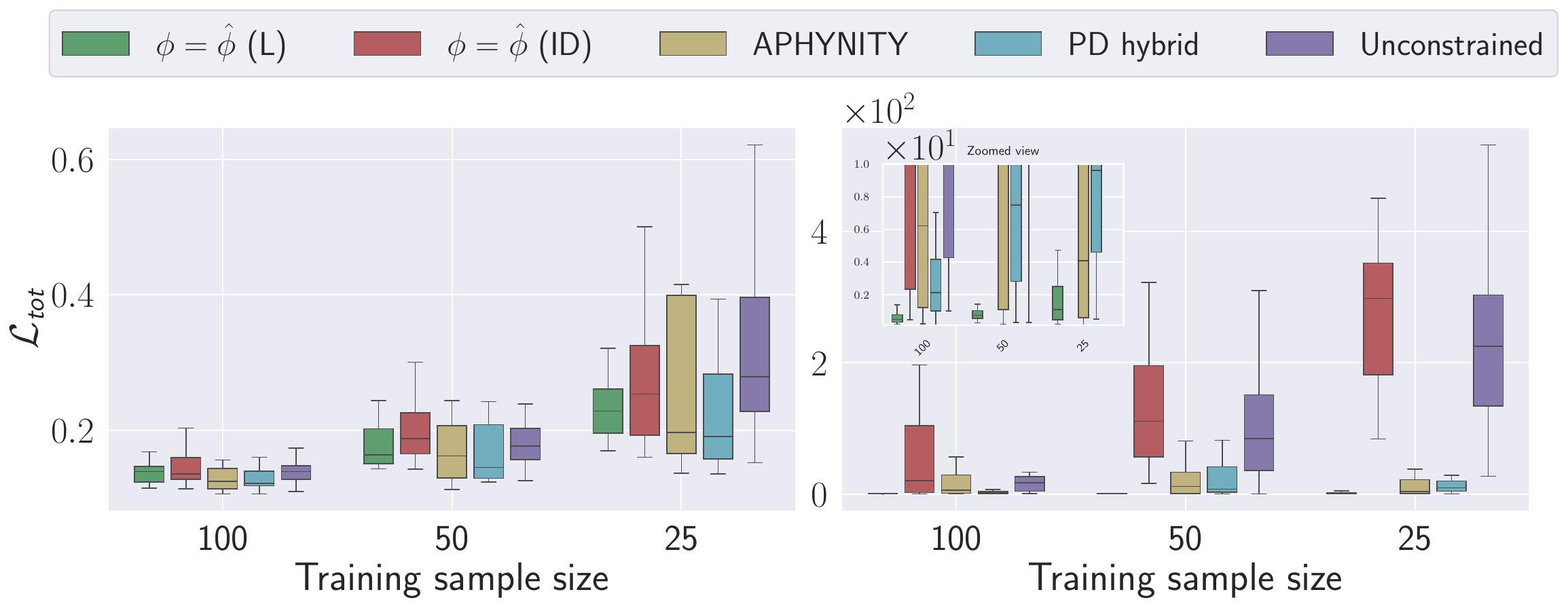}
    \caption{Evolution of $\mathcal{L}_{\textit{tot}}$ w.r.t. the training sample size, on the PHALK dataset, constraining $\textit{PD}_N(h^{\theta}, \text{alkalinity})$, in the interpolation setting (left) and extrapolation setting (right). Each boxplot summarizes the results on a test set, over 20 different dataset splits.}
    \label{fig:pdp_phalk_mse_interpolation_extrapolation}
\end{figure}

\paragraph{Sample size effect}
We observe in Figure \ref{fig:pdp_phalk_mse_interpolation_extrapolation} (left) that hybrid methods reach better performance than unconstrained networks, for all training sizes. 

\paragraph{Test-time domain shift}
We observe in Figure \ref{fig:pdp_phalk_mse_interpolation_extrapolation} (right) that hybrid methods reach worse performance with out-of-distribution samples, yet performing better than in-distribution constraining and unconstrained models. For this problem, we observe that linspace constraining vastly outperforms all the other methods.

\begin{figure}[ht]
    \centering
    \includegraphics[width=\linewidth]{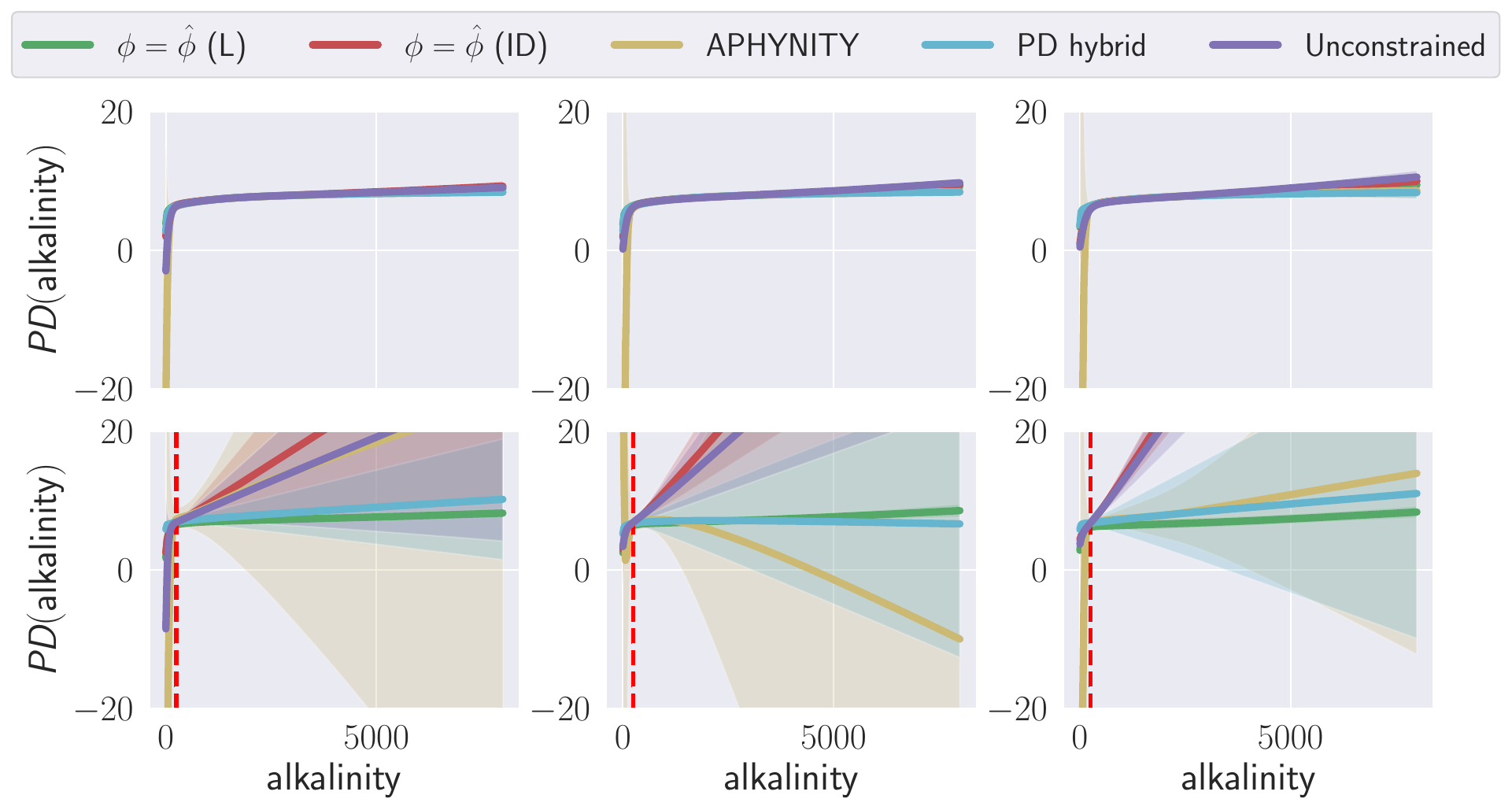}
    \caption{The figure displays the mean PD estimations for $\textit{PD}_N(h^{\theta}, \text{alkalinity})$ obtained with each method, for the PHALK dataset, in the interpolation (top) and extrapolation settings (bottom). All methods have been trained respectively on 100 samples (left), 50 samples (middle), 25 samples (right). Mean and standard deviations are computed over 20 different dataset splits. The dotted red lines represent the boundaries of the interpolation range.}
    \label{fig:pdp_phalk_all_sizes_pd_comparison}
\end{figure}

We represent in Figure \ref{fig:pdp_phalk_all_sizes_pd_comparison} the mean PD estimations, for all methods and training sizes. We notice that in the interpolation setting (top), the PD estimations all follow a logarithmic trend. However, in the extrapolation setting (bottom), the PD estimations of APHYNITY show stability problems, but generally do not look logarithmic. On the contrary, PD hybrid modeling suffers less than APHYNITY, as its PD estimation looks close to that of linspace constraining.

\clearpage

\section{Constraining with mis-specified knowledge}
\label{app:misspecified_prior}
In this section, we study the performance of our constraining methods in the presence of mis-specified prior knowledge, thereby assuming that the expert poorly models the true partial dependence function. For that purpose, we rely on models whose hypothesis spaces do not include the true PD model but respect similar properties.

For both experiments, we fix the architecture of the neural network to three layers of increasing hidden size (32, 64 and 128), with a fixed training batch size of 16 samples for the Friedman problem, and 5 for the concrete compressive strength problem. We optimize the learning rate $\tau_2$ in $\{\num{5e-2}, \num{5e-3}, \num{5e-4}\}$ and the coefficient $\lambda$ in $\{0.25, 0.5, 0.75\}$, selecting the values that minimize $\mathcal{L}_{\textit{tot}}$ on the validation set. 

\subsection{Friedman problem}
Let us recall the problem definition:
\begin{align*}
    y = \psi_{0} \sin(&\psi_{1} x_0 x_1) + \psi_{2} (x_2 - \psi_{3})^2 + \psi_{4} x_3 + \psi_{5} x_4 + \sum_{j=5}^{9}0 x_j + \varepsilon,
\end{align*}
where $x_j \sim \mathcal{U}(0, 1), j=0, \dots 9$, $\psi = [10, \pi, 20, 0.5, 10, 5]$ and $\varepsilon \sim \mathcal{N}(0, 1)$ \citep{friedman1983multidimensional}.

In this setting, we only assume the prior knowledge to be a periodic function 
\begin{align}
\label{eq:friedman_imperfect_equation}
    h_k^{\phi}(x_0, x_1; \phi, \phi_{\sin}, \phi_{\cos}) = \phi_0 [&\sum_{l=1}^{L} \phi_{\sin}^{(l)} \sin(2\pi l x_0 x_1)  + \phi_{\cos}^{(l)} \cos(2\pi l x_0 x_1)] + \phi_1
\end{align}
through a Fourier series with period 1, where $\phi_{\sin} \in \R^{L}, \phi_{\cos} \in \R^L$. We denote linspace constraining with this imperfect setting as $\phi = \hat{\phi}$ (L - Imperf.), and in-distribution constraining as $\phi = \hat{\phi}$ (ID - Imperf.).

Let us remind that the true PD function should be 
\begin{align*}
    h_k^{\phi^*}(x_0, x_1) &= \; 
        1 \cdot[10 \sin(\pi x_0 x_1)] + C_1,
\end{align*}
where
\begin{align*}
    C_1 &= \mathbb{E}_{\x_{-k}}\left[\psi_{2} (x_2 - \psi_{3})^2 + \psi_{4} x_3 + \psi_{5} x_4\right].
\end{align*}
However, given that $h_k^{\phi}(x_0, x_1)$ here solely relies on basis terms with frequencies that are integer multiples of $2\pi$, it is therefore impossible to recover exactly the true PD function with $h_k^\phi$, hence the knowledge is imperfect by construction, but it is still able to provide a decent representation. For this experiment, we set $L = 3$.

\begin{figure}[ht]
    \centering
    \includegraphics[width=\linewidth]{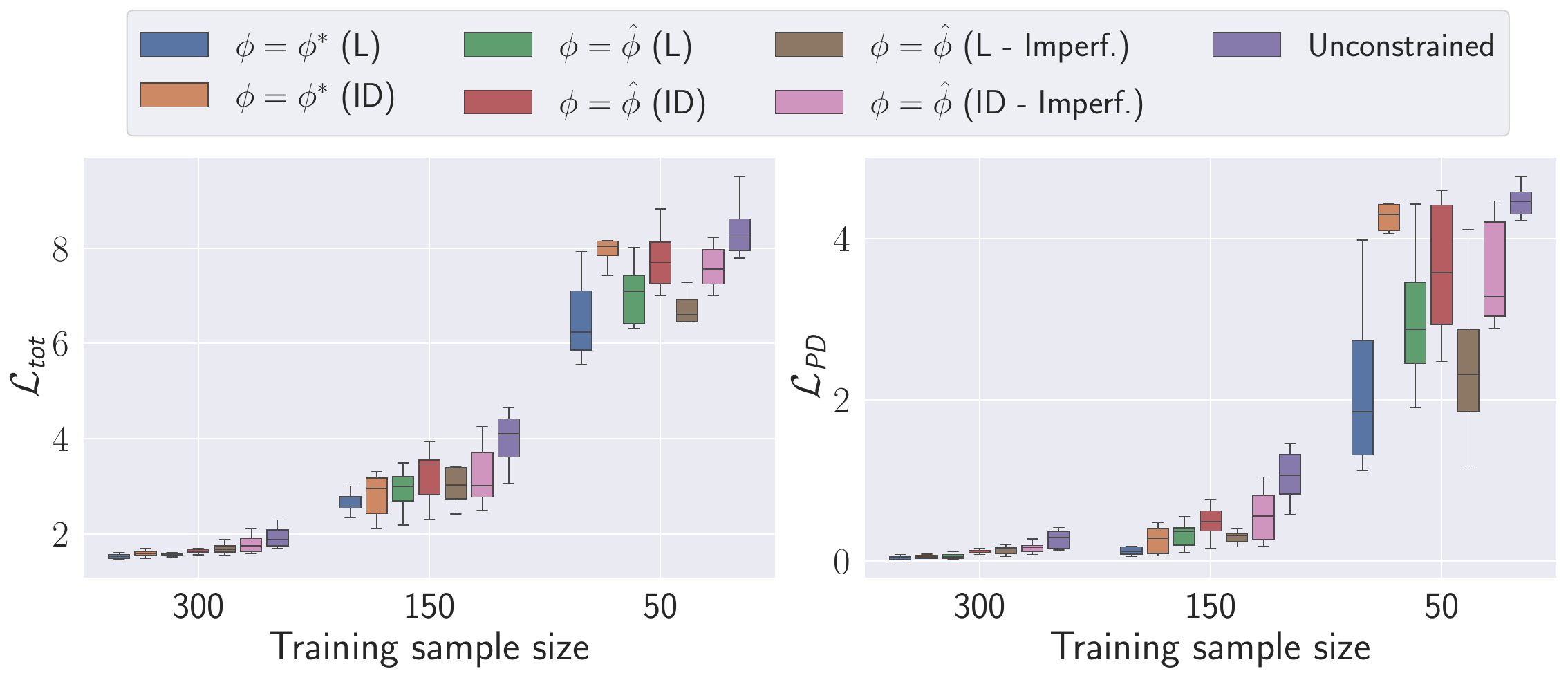}
    \caption{Evolution of $\mathcal{L}_{\textit{tot}}$ (left) and $\mathcal{L}_{\textit{PD}}$ (right) w.r.t. the training sample size, on the Friedman problem, constraining $\textit{PD}_N(h^{\theta}, x_0, x_1)$. Each boxplot summarizes the results on the test set, over 10 different training initializations.}
    \label{fig:imperfect_friedman_first_mse_pd_mse}
\end{figure}

\paragraph{Sample size effect}
We observe in Figure \ref{fig:imperfect_friedman_first_mse_pd_mse} that decreasing the learning sample size degrades both measures of performance. Nevertheless, constraining the PD of the model with such imperfect knowledge still improves both measures compared to unconstrained training, for all training sizes.

\begin{figure}[ht]
    \centering
    \includegraphics[width=\linewidth]{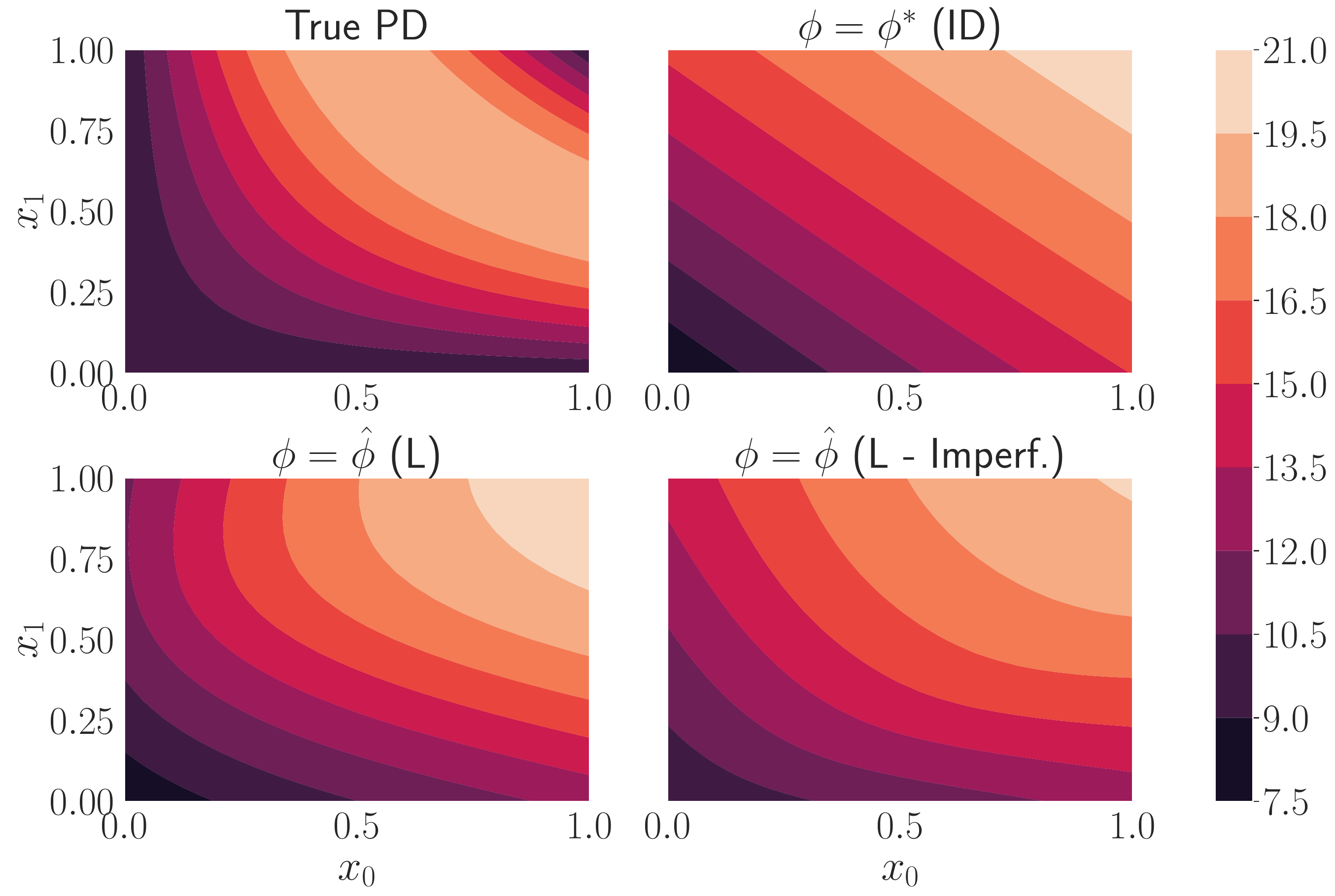}
    \caption{Examples of PD estimation for $\textit{PD}_N(h^{\theta}, x_0, x_1)$. The figure displays the true PD (upper left), the PD after in-distribution constraining on the ground truth PD (upper right), the PD after linspace constraining on the approximated prior (bottom left), and the PD after linspace constraining with imperfect prior knowledge (bottom right). All have been trained on 50 samples, in the interpolation setting.}
    \label{fig:imperfect_friedman_pd_first_50_samples_interpolation}
\end{figure}

Even with mis-specified prior knowledge, linspace constraining (brown boxes) performs on par with in-distribution constraining on true PD samples (orange boxes), and even outperforms the latter when data becomes very scarce (50 samples). This can be verified by looking at Figure \ref{fig:imperfect_friedman_pd_first_50_samples_interpolation}, which highlights the difference in the PD estimates. In general, constraining with imperfect prior knowledge (brown and pink boxes) performs as well as constraining on the true PD with approximated parameters (green and red boxes), but we notice that the former actually outperforms the latter for 50 samples, which is due to the fact that the latter does not need to approximate the frequency parameter of the sine function, making it less prone to overfitting.

\paragraph{Test-time domain shift}
We measure performance under test-time domain shift in identical conditions, i.e. samples for which $x_0 > 0.75$ are rejected from the training and validation sets.

\begin{figure}[ht]
    \centering
    \includegraphics[width=\linewidth]{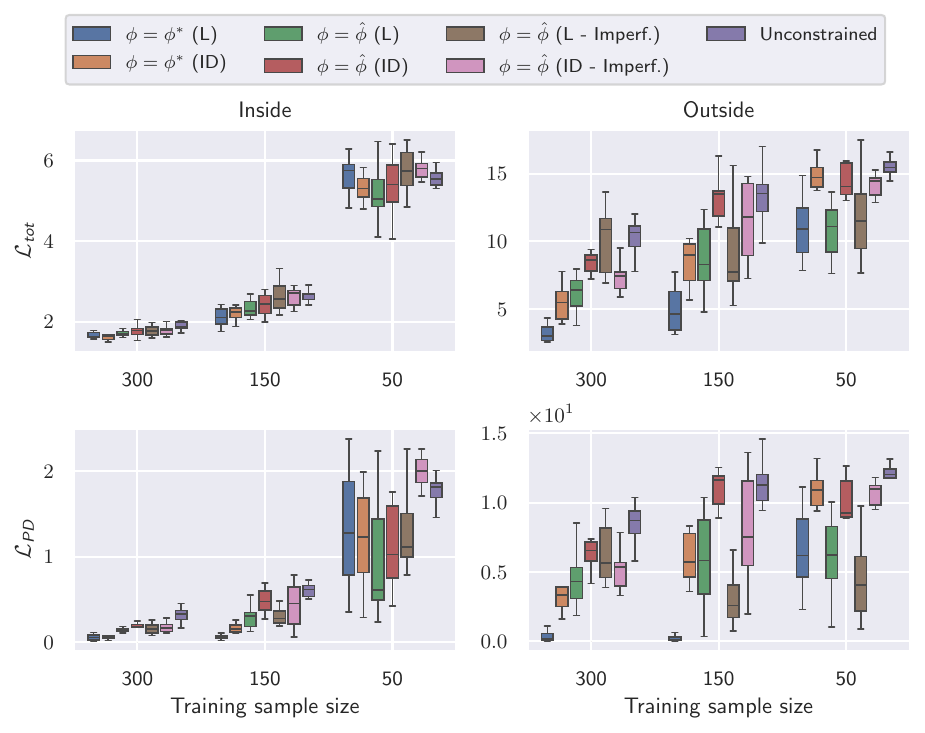}
    \caption{Evolution of $\mathcal{L}_{\textit{tot}}$ (top plots) and $\mathcal{L}_{\textit{PD}}$ (bottom plots) w.r.t. the training sample size, on the Friedman problem, constraining $\textit{PD}_N(h^{\theta}, x_0, x_1)$. \textit{Inside} metrics are computed on samples for which $x_0 \leq 0.75$ (left plots), while \textit{Outside} metrics are computed on samples where $x_0 > 0.75$ (right plots). Each boxplot summarizes the results on the test set, over 10 different training initializations.}
    \label{fig:imperfect_friedman_first_inside_outside}
\end{figure}

We can observe in Figure \ref{fig:imperfect_friedman_first_inside_outside} that, for both metrics, PD steering helps to mitigate the impact of test-time domain shift, and much more when constraining on evenly-spaced points. As for the interpolation setting, we can notice that our method is robust to imperfect knowledge in the sense of \eqref{eq:friedman_imperfect_equation}.

\subsection{Concrete compressive strength prediction}
As a reminder, the dataset is composed of samples relating characteristics of concrete components with the associated compressive strength. Instead of assuming the relationship between age and compressive strength to be logarithmic, we model the latter as a square root
\begin{align*}
    h_k(\text{age}; \phi) = \phi_0\sqrt{\text{age}} + \phi_1.
\end{align*}
The motivation behind this choice is to assume that prior knowledge is limited to a strictly increasing concave function, hence the expert wrongly modeled the PD with a square root function.

\begin{figure}[ht]
    \centering
    \includegraphics[width=\linewidth]{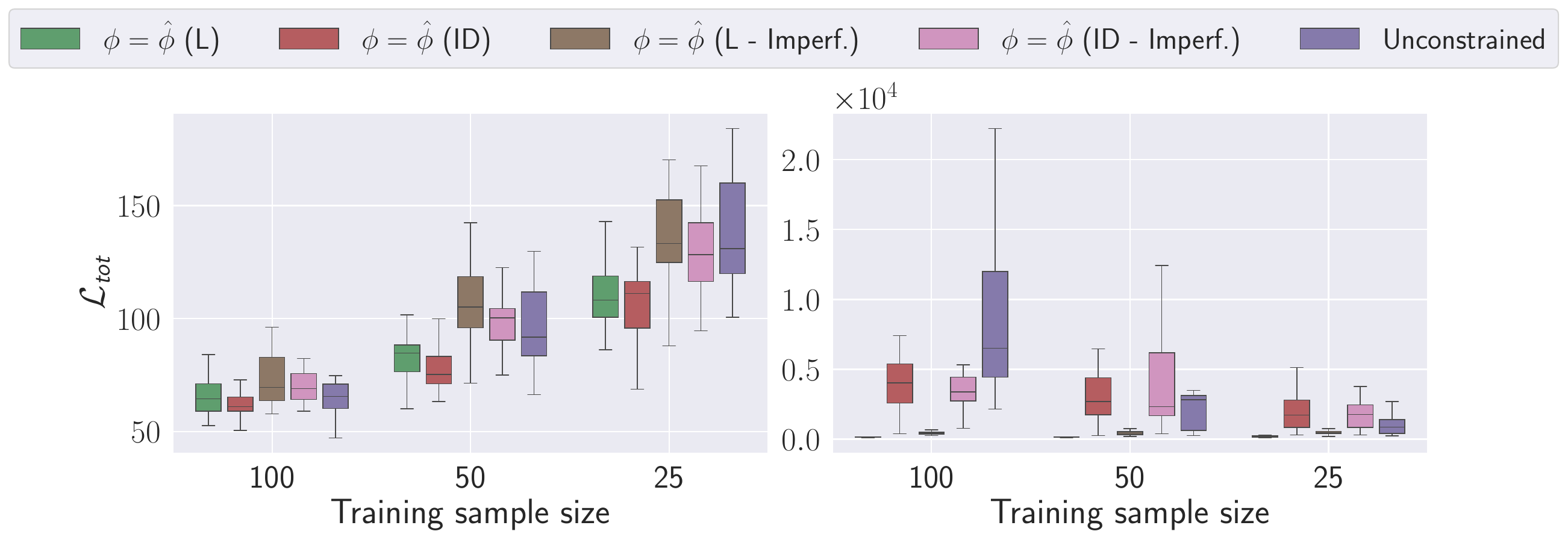}
    \caption{Evolution of $\mathcal{L}_{\textit{tot}}$ w.r.t. the training sample size, on the concrete problem, constraining $\textit{PD}_N(h^{\theta}, \text{age})$, in the interpolation setting (left) and extrapolation setting (right). Each boxplot summarizes the results on a test set, over 20 different dataset splits.}
    \label{fig:imperfect_concrete_mse_interpolation_extrapolation}
\end{figure}

\paragraph{Sample size effect}
We observe in Figure \ref{fig:imperfect_concrete_mse_interpolation_extrapolation} (left) that when constraining with imperfect knowledge, the performance of the resulting models is no better than that of the unconstrained models, which was not the case when assuming a logarithmic PD. This behavior actually makes sense, since we are trying to constrain the PD of the neural network to follow an incorrect shape.

Contrary to the previous problem where periodicity was the only assumption, the source of mis-specification in this case is tougher as there is no way for a square root function to yield a reasonable approximation of a logarithmic function.

\paragraph{Test-time domain shift}
Identically to the original extrapolation setting, we have removed from training and validation sets samples whose age value was larger than 15 days.

We observe in Figure \ref{fig:imperfect_concrete_mse_interpolation_extrapolation} (right) that, in this case, the performance of models constrained with imperfect prior knowledge is actually very close to that of models constrained with the logarithmic PD. As was the case, linspace constraining outperforms both in-distribution constraining and unconstrained training, for all training sizes, and reaches performance close to that obtained in the interpolation setting, even though prior knowledge is fundamentally mis-specified.

\begin{figure}[ht]
    \centering
    \includegraphics[width=\linewidth]{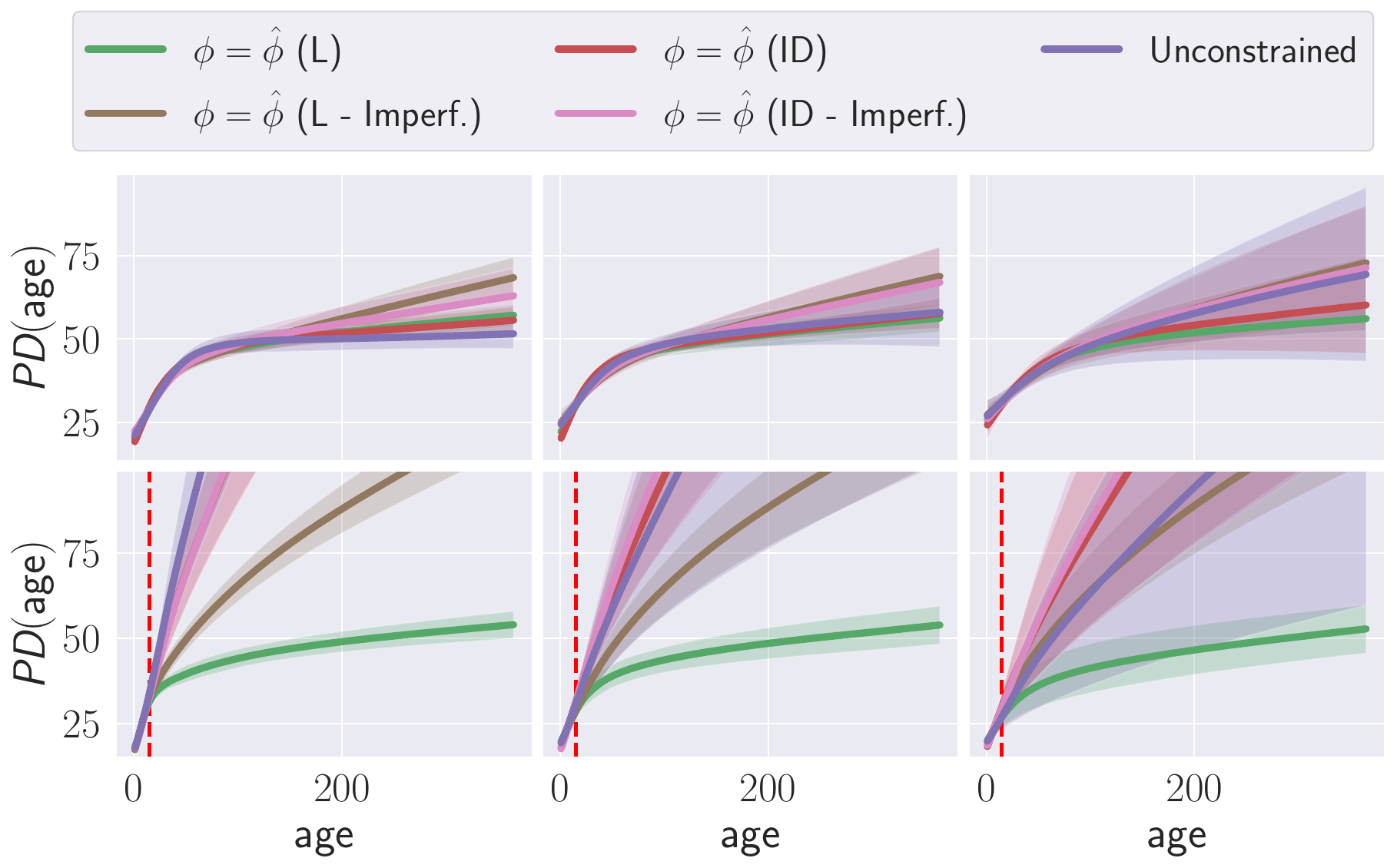}
    \caption{The figure displays the mean PD estimations for $\textit{PD}_N(h^{\theta}, \text{age})$ obtained with each method, for the compressive strength problem, in the interpolation (top) and extrapolation settings (bottom). All methods have been trained respectively on 100 samples (left), 50 samples (middle), 25 samples (right). Mean and standard deviations are computed over 20 different dataset splits. The dotted red lines represent the boundaries of the interpolation range.}
    \label{fig:imperfect_concrete_all_sizes_pd_comparison}
\end{figure}

In Figure \ref{fig:imperfect_concrete_all_sizes_pd_comparison}, we provide the mean PD estimations for all methods, both in the interpolation (top) and extrapolation (bottom) settings. In the interpolation setting, we can observe that the PD estimations of models constrained with imperfect knowledge indeed follow a square root trend, as opposed to the logarithmic trend observed for the other models. Interestingly, when data becomes very scarce (25 samples), we notice that those of unconstrained models tend to approach the square root trend.

In the extrapolation setting, the situation is different. Indeed, unconstrained models and both in-distribution constraining (red and pink curves) fail to capture the logarithmic shape outside the training domain, which confirms the poor performance that was observed in Figure \ref{fig:imperfect_concrete_mse_interpolation_extrapolation} (right). Even more, we notice that the PD estimations of both in-distribution constraining are very close, which makes sense given the resemblance of the logarithm and square root functions for small inputs. Despite imperfect prior knowledge, linspace constraining helps to approach a shape reasonably close to the desired PD, thereby reducing the prediction error, as was observed in Figure \ref{fig:imperfect_concrete_mse_interpolation_extrapolation} (right). With very few data samples (25), we can observe that the PD of unconstrained models approaches that of linspace-constrained models (brown curve), although it still seems to diverge for larger values of age, which explains the gap in prediction error.

\clearpage

\section{List of variables}
\subsection{Concrete compressive strengh dataset}
\label{app:concrete_variables_description}

\begin{table}[h]
    \caption{Variables used for the concrete compressive strength dataset. The variable indicated in bold type is the one for which PD prior knowledge is assumed ($x_k$).}
    \centering
    \begin{tabular}{ll}
    \toprule
    Name & Description\\
    \midrule
    Cement & Amount of cement in the mixture $\operatorname{[kg/m^3]}$\\
    Blast Furnace Slag & Amount of blast furnace slag in the mixture $\operatorname{[kg/m^3]}$ \\
    Fly Ash & Amount of fly ash in the mixture $\operatorname{[kg/m^3]}$ \\
    Water & Amount of water in the mixture $\operatorname{[kg/m^3]}$ \\
    Superplasticizer & Amount of superplasticizer in the mixture $\operatorname{[kg/m^3]}$ \\
    Coarse Aggregate & Amount of coarse aggregate in the mixture $\operatorname{[kg/m^3]}$ \\
    Fine Aggregate  & Amount of fine aggregate in the mixture $\operatorname{[kg/m^3]}$ \\
    \textbf{Age} & \textbf{Day (1-365)} \\
    \bottomrule
    \end{tabular}
\end{table}

\subsection{PHALK dataset}
\label{app:phalk_variables_description}

\begin{table}[h]
    \caption{Variables used in the PHALK dataset. The variable indicated in bold type is the one for which PD prior knowledge is assumed ($x_k$).}
    \centering
    \begin{tabular}{ll}
    \toprule
    Name & Description\\
    \midrule
    Latitude & Latitude of the basin centroid in WGS84\\
    Longitude & Longitude of the basin centroid in WGS84 \\
    \textbf{Alkalinity} & \textbf{Estimated alkalinity values} $\operatorname{[\mu eq/l]}$ \\
    $\textit{XX}_{su}^U$ & \% of unconsolidated sediments within the total watershed upstream of the sub-basin \\
    $\textit{XX}_{ss}^U$ & \% of siliclastic sedimentary rocks within the total watershed
upstream of the sub-basin \\
    $\textit{XX}_{sm}^U$ & \% of mixed sedimentary rocks within the total watershed
upstream of the sub-basin \\
    $\textit{XX}_{sc}^U$ & \% of carbonate sedimentary rocks within the total watershed
upstream of the sub-basin \\
    $\textit{XX}_{py}^U$ & \% of pyroclastics within the total watershed upstream of the sub-basin \\
    $\textit{XX}_{ev}^U$ & \% of evaporates within the total watershed upstream of the sub-basin \\
    $\textit{XX}_{mt}^U$ & \% of metamorphic rocks within the total watershed upstream of
the sub-basin \\
    $\textit{XX}_{pa}^U$ & \% of acid plutonic rocks within the total watershed upstream of
the sub-basin \\
    $\textit{XX}_{pi}^U$ & \% of intermediate plutonic rocks within the total watershed
upstream of the sub-basin \\
    $\textit{XX}_{pb}^U$ & \% of basic plutonic rocks within the total watershed upstream of
the sub-basin \\
    $\textit{XX}_{va}^U$ & \% of acid volcanic rocks within the total watershed upstream of
the sub-basin \\
    $\textit{XX}_{vi}^U$ & \% of intermediate volcanic rocks within the total watershed
upstream of the sub-basin \\
    $\textit{XX}_{vb}^U$ & \% of basic volcanic rocks within the total watershed upstream of
the sub-basin \\
    $\textit{XX}_{ig}^U$ & \% of ice and glaciers within the total watershed upstream of the
sub-basin \\
    $\textit{XX}_{wb}^U$ & \% of water bodies within the total watershed upstream of the sub-basin \\
    Slope degree & Terrain slope (º) within total watershed upstream of sub-basin \\
    Water surface & 1 if basin with water surface \\
    \bottomrule
    \end{tabular}
\end{table}

\end{document}